\documentclass[sn-mathphys-num]{sn-jnl}
\usepackage{graphicx}%
\usepackage{multirow}%
\usepackage{amsmath,amssymb,amsfonts}%
\usepackage{amsthm}%
\usepackage{mathrsfs}%
\usepackage[title]{appendix}%
\usepackage{xcolor}%
\usepackage{textcomp}%
\usepackage{manyfoot}%
\usepackage{booktabs}%
\usepackage{algorithm}%
\usepackage{listings}%
\usepackage{algorithmic}
\usepackage{subfigure}

\begin{document}

\title[Article Title]{Enhancing Missing Data Imputation through Combined Bipartite Graph and Complete Directed Graph}

\author[1]{\fnm{Zhaoyang} \sur{Zhang}}\email{10215001426@stu.ecnu.edu.cn}

\author[2]{\fnm{Hongtu} \sur{Zhu}}\email{htzhu@email.unc.edu}
\author*[1]{\fnm{Ziqi} \sur{Chen}}\email{zqchen@fem.ecnu.edu.cn}
\author[1]{\fnm{Yingjie} \sur{Zhang}}\email{2427785647@qq.com}
\author[3]{\fnm{Hai} \sur{Shu}}\email{hai.shu@nuy.edu}

\affil*[1]{\orgdiv{School of Statistics}, \orgname{East China Normal University}, \orgaddress{\city{Shanghai}, \country{China}}}

\affil[2]{\orgdiv{Departments of Biostatistics, Statistics, Computer Science and Genetics}, \orgname{The University of North Carolina at Chapel Hill}, \orgaddress{\city{Chapel Hill}, \country{USA}}}

\affil[3]{\orgdiv{Department of Biostatistics, School of Global Public Health}, \orgname{New York University}, \orgaddress{\city{New York}, \country{USA}}}

\abstract{In this paper, we aim to address a significant challenge in the field of missing data imputation: identifying and leveraging the interdependencies among features to enhance missing data imputation for tabular data. We introduce a novel framework named the Bipartite and Complete Directed Graph Neural Network (BCGNN). Within BCGNN, observations and features are differentiated as two distinct node types, and the values of observed features are converted into attributed edges linking them. The bipartite segment of our framework inductively learns embedding representations for nodes, efficiently utilizing the comprehensive information encapsulated in the attributed edges. In parallel, the complete directed graph segment adeptly outlines and communicates the complex interdependencies among features. When compared to contemporary leading imputation methodologies, BCGNN consistently outperforms them, achieving a noteworthy average reduction of 15\% in mean absolute error for feature imputation tasks under different missing mechanisms. Our extensive experimental investigation confirms that an in-depth grasp of the interdependence structure substantially enhances the model's feature embedding ability. We also highlight the model's superior performance in label prediction tasks involving missing data, and its formidable ability to generalize to unseen data points.}

\keywords{Graph Neural Network, Missing Data, Missing Mechanism}

\maketitle

\section{Introduction}
\label{submission}
Incomplete data is a common occurrence in fields such as clinical research, finance, and economics and survey research \cite{ibrahim2009missing,ibrahim2005missing,brick1996handling,wooldridge2007inverse,sterne2009multiple}. Tackling incomplete data for downstream learning tasks requires a comprehensive strategy focused on preserving data integrity and minimizing bias. A prevalent approach is feature imputation, where missing values are estimated using statistical techniques such as regression imputation, as well as more advanced methods, including machine learning algorithms and deep learning models, which capture the underlying relationships in the data \cite{little2019statistical,loh2011classification,smieja2018processing,you2020handling,kyono2021miracle}.

\begin{figure}[t] 
	\centering 
	\includegraphics[height=3cm,width=13cm]{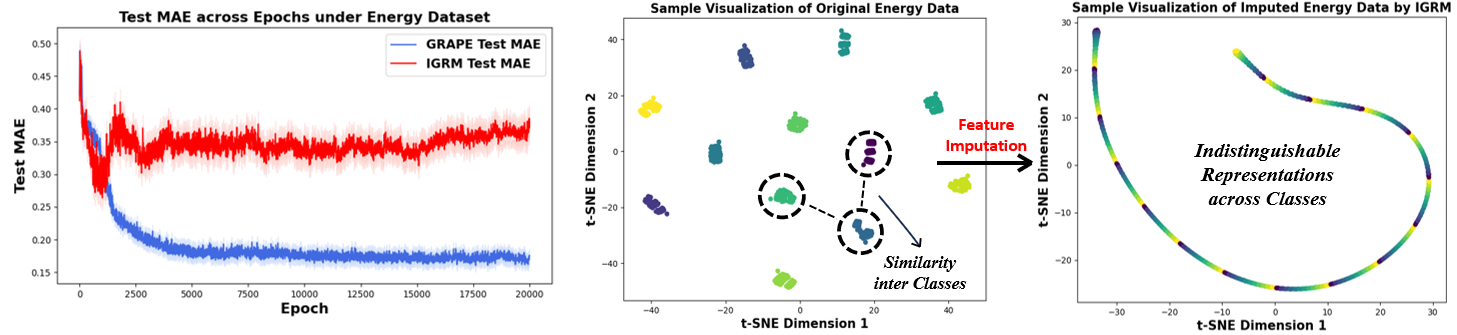}
	\caption{Comparison of GRAPE and IGRM on the Energy dataset in terms of test MAE in feature imputation (\textbf{Left}); Visualizations of the original and the imputed (by IGRM) Energy datasets using t-SNE for dimensionality reduction (\textbf{Middle} and \textbf{Right}). The poor performance of IGRM can be attributed to erroneously identifying similarities among samples from different classes, leading to indistinguishable sample representations across classes.}
	\vspace{-3mm}
	\label{comparison}
\end{figure}

When addressing missing data, it is common to classify data structures into two main categories: graph data and  tabular data. For graph data, where the connectivity among samples is considered prior knowledge, numerous studies enhance feature imputation based on the known adjacency information among samples, subsequently facilitating downstream tasks such as node classification and link prediction \cite{zhang2023missing,um2023confidence,gupta2023grafenne}. For tabular data, the focus is on the accuracy of the feature imputation and label prediction  \cite{you2020handling,zhong2023data,telyatnikov2023egg}. 
Unlike graph data, samples in tabular data are independent of each other. This independence renders most feature imputation methods used in graph data inapplicable to tabular data due to the absence of  connecting edges among sample nodes. Nevertheless, some existing methods for feature imputation in tabular data concentrate on capturing the relationships among samples. These methods primarily aim to identify similarities among samples to enhance feature imputation \cite{zhong2023data,gupta2023grafenne}.
However, solely focusing on the similarities and connectivity among independent samples in tabular data can be detrimental. For example,   Figure \ref{comparison} (left panel) reveals that IGRM, a variant of GRAPE that considers similarities among samples, performs  worse on the Energy dataset compared to the original GRAPE \cite{zhong2023data,you2020handling}. Moreover, Figure \ref{comparison} (middle and right panels)   illustrates the limitations of IGRM, which only captures similarities among samples when handling missing data.

In this paper, we introduce a cutting-edge framework known as Bipartite and Complete directed Graph Neural Network (BCGNN), which synergizes the structures of bipartite and complete directed graphs to proficiently address the feature imputation task for tabular data. This approach is rooted in inductive learning and a deep comprehension of feature interdependencies. 

In BCGNN, the bipartite graph and complete directed graph serve distinct functions. In the bipartite graph, observation nodes and feature nodes aggregate messages from neighboring nodes and corresponding edges to learn their embedding representations. Meanwhile, in the complete directed graph, we innovatively implement an element-wise attention mechanism on the embeddings of feature nodes and introduce the signs of estimated Spearman correlation coefficients to learn and express the interdependence structure among features. The BCGNN framework is constructed by stacking multiple layers of the union of these two subgraphs. At each layer, we perform complete message aggregation and embedding updates on each node, followed by the updates of edge embeddings. Ultimately, the task of feature imputation can be accomplished by integrating the embeddings of corresponding feature nodes and observation nodes at the final layer.

\subsection{Related Work}
\label{Relatedwork}
\textbf{Statistical methods}. Classical statistics techniques, such as the Expectation Maximization (EM) algorithm and Multivariate Imputation by Chained Equations (MICE), which rely on distributional assumptions \cite{burgette2010multiple, van2011mice,little2019statistical,zhu2001local,ibrahim2005missing}, have been successful in imputation tasks. Additionally, k-nearest neighbors (KNN) imputation \cite{kim2004reuse} and matrix completion techniques \cite{srebro2004maximum, cai2010singular, hastie2015matrix} have shown effectiveness. However, EM and MICE typically assume a parametric density function for data distribution, which poses limitations when dealing with diverse missing mechanisms and mixed data types. Both MICE and KNN suffer from scalability issues and lack flexibility for downstream tasks.  Matrix completion methods also find it challenging to accommodate mixed data types and become computationally intensive for imputing new observations after model training. 

\textbf{Machine and deep learning based methods}. Recent deep learning models applied to feature imputation exhibit notable limitations. Generative Adversarial Imputation Nets (GAIN) \cite{yoon2018gain} struggle with missing not at random (MNAR) scenarios. Denoising Autoencoder (DAE) models \cite{vincent2008extracting,gondara2017multiple} are limited by their reliance on a single observation, failing to capture complex inter-observation interactions. Optimal Transport (OT) \cite{muzellec2020missing} assumes identical distributions for two randomly extracted batches from the same dataset and employs optimal transport distance as the training loss. MIRACLE  \cite{kyono2021miracle}, using causal graphs, faces challenges due to its dependence on restrictive assumptions, limiting its flexibility. 

\textbf{Graph-based framework}. Numerous  advanced graph-based methods have been proposed for data imputation tasks. Some of these methods are developed for matrix completion and demonstrate excellent performance on large matrices \cite{berg2017graph, zhang2019inductive, li2021deconvolutional, zhang2023missing}. However, their assumptions about feature types limit their applicability to mixed discrete and continuous features. Some  Graph Neural Network (GNN)-based models,  which employ diversified interaction and information propagation strategies among sample nodes based on prior known adjacency information, successfully enhance feature imputation in graph data for downstream tasks like node classification and link prediction \cite{zhang2023missing,um2023confidence,gupta2023grafenne}. However, they prove ineffective in tabular data due to the absence of prior topological connectivity among samples. GRAPE \cite{you2020handling} showcases success by leveraging a bipartite graph for feature imputation. Following GRAPE, IGRM \cite{zhong2023data} goes a step further by establishing a friend network among sample nodes to enhance feature imputation by capturing similarities among samples. Meanwhile, GINN~\cite{spinelli2020missing} utilizes the Euclidean distance of observed data to construct a graph and then reconstructs the tabular data. EGG-GAE \cite{telyatnikov2023egg} learns latent topological structures among sample nodes to assist feature imputation for various downstream tasks. 

\subsection{Our contributions}

It is notable that the current advanced methods for enhancing feature imputation overlook the interdependencies among features. The intricate interdependence among features provides crucial insights for imputing missing data, particularly in scenarios with high missing rates and diverse missing mechanisms. Our work stands as the pioneering effort to capture the interdependencies among features by integrating a complete directed graph into the graph neural network framework. This integration introduces three significant innovations in our BCGNN model:

First, our BCGNN capture the interdependence among features by combining a complete directed graph with a bipartite graph. This integration enables the model to learn node representations inductively and maps out the interdependence between features in a cohesive structure.

Second, our BCGNN employs a unique element-wise additive attention mechanism within the complete directed graph. This allows for a detailed learning of how one feature depends on the attributes of another, a method totally distinct from the usual node-level or edge-level attention in GNNs \cite{velivckovic2017graph,chen2021edge}. By also incorporating the sign of Spearman correlation coefficients, the model gains a refined understanding of feature relationships, enabling more precise imputation.

Lastly, to prevent overfitting, BCGNN uses the DropEdge method \cite{rong2019dropedge} across both subgraphs. For enhancing generalization, the framework includes an AttentionDrop feature within the element-wise attention mechanism, preventing undue emphasis on particular segments of the embeddings. 

\begin{figure}[t] 
	\centering 
	\includegraphics[height=4.5cm,width=13.3cm]{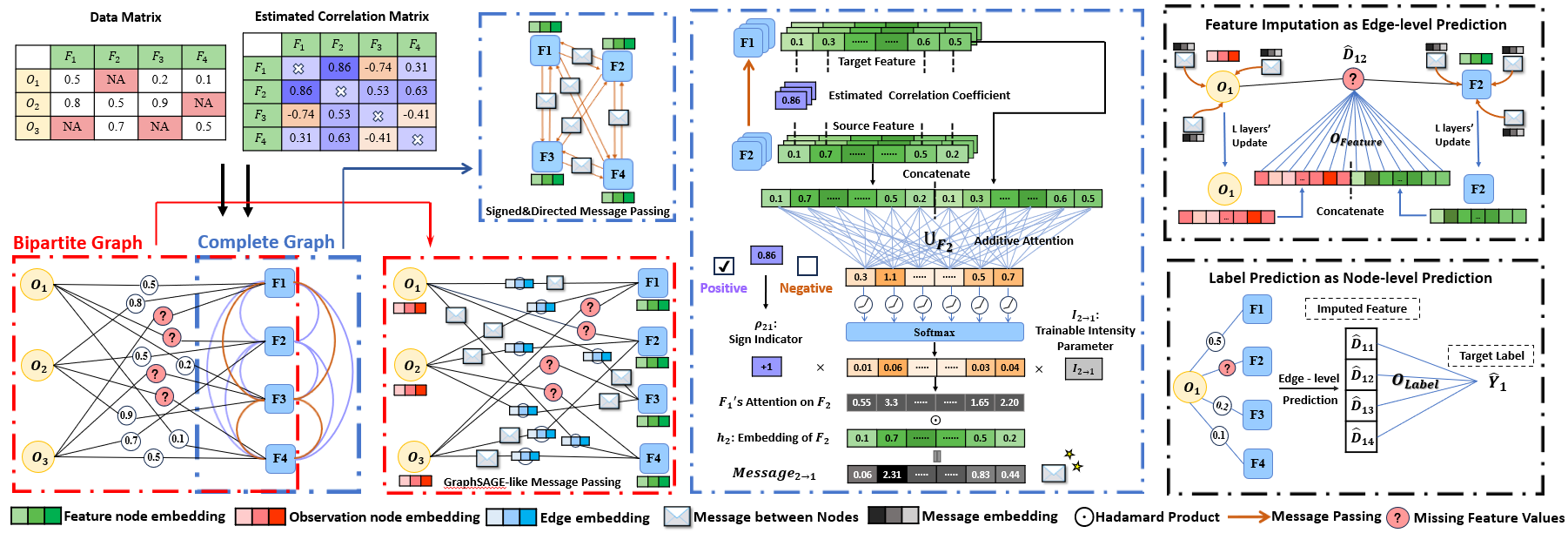}
	\caption{Flowchart of our BCGNN method. BCGNN consists of a bipartite graph and a complete directed graph constructed from the data matrix and correlation coefficient matrix of features. The working mechanisms of the bipartite graph (\textbf{Red Dot-Dashed Box}), the complete directed graph (\textbf{Blue Dot-Dashed Box}) and the union of two subgraphs are elaborated in detail in Sections \ref{bipartite}, \ref{Complete} and \ref{union}, respectively. With the constructed graph, the feature imputation problem and the label prediction problem are treated as edge-level and node-level prediction tasks, respectively (\textbf{Black Dot-Dashed Box}).}
	\vspace{-3mm}
	\label{BCGNN}
\end{figure}

\section{Model Architecture and Algorithm}
\subsection{Problem Definition} 

Let $\mathbf{D} \in \mathbb{R}^{n \times m}$ be a data matrix composed of $n$ observational samples and $m$ features. The $j$-th feature of the $i$-th observation sample is denoted as $D_{ij}$. Let $\mathbf{D}_{i\cdot}$ be the $i$-th row of $\mathbf{D}$. We use a masking matrix $\mathbf{M}=(M_{ij}) \in \{0,1\}^{n \times m}$ to identify the missing data in data matrix. Specifically,  $M_{ij} = 1$ indicates that $D_{ij}$ is observed, while $M_{ij} = 0$ indicates that it is missing. Let $\mathbf{Y}=(Y_i)\in \mathbb{R}^{n}$ denote the labels for the $n$ samples, we use a corresponding masking vector $\mathbf{M}^y=(M^y_i) \in \{0,1\}^{n}$ to identify the missing values in $\mathbf{Y}$. We consider feature imputation and label prediction task, i.e, imputing missing feature values $D_{ij}$ at $M_{ij} = 0$ and predicting labels $Y_i$ at $M^y_i=0$.
\subsection{Missing Data Problem as a Union of Two Graphs}
\label{problemdefinition}
Our BCGNN  integrates node embeddings, edge embeddings, and a specially designed message-passing mechanism, effectively tackling challenges in missing data imputation. Specifically, BCGNN transforms a data matrix with missing data into a union of two subgraphs as illustrated in Figure \ref{BCGNN}. One subgraph, denoted as a bipartite graph $\mathcal{G}_B$, consists of observation nodes, feature nodes, and the edges connecting these two types of nodes. Meanwhile, the other subgraph $\mathcal{G}_C$ is a complete directed graph  formed exclusively among the feature nodes.
The union graph can then be represented as $\mathcal{G} = \mathcal{G}_B \cup \mathcal{G}_C$. Consequently, the data matrix $\mathbf{D}$ and the masking matrix $\mathbf{M}$ can be naturally represented as a graph $\mathcal{G} = (\mathcal{V}, \mathcal{E})$, where $\mathcal{V}$ includes two types of nodes $\mathcal{V} = \mathcal{V}_F \cup \mathcal{V}_O$, with $\mathcal{V}_O = \{u_1, u_2, \ldots, u_n\}$ representing observation nodes and $\mathcal{V}_F = \{v_1, v_2, \ldots, v_m\}$ representing feature nodes. 
$\mathcal{E}$ contains two types of edges $\mathcal{E}=\mathcal{E}_{OF}\cup{\mathcal{E}_{FF}}$, where $\mathcal{E}_{OF} = \{e_{u_iv_j} |  u_i \in \mathcal{V}_O, v_j \in \mathcal{V}_F, M_{ij} = 1\}$ represents edges connecting feature nodes with observation nodes, and $\mathcal{E}_{FF} = \{(e_{v_i\rightarrow v_{j}}, e_{v_{j}\rightarrow v_i}) | v_{i}, v_j \in \mathcal{V}_F,v_i\neq{v_j}\}$  denotes a set of directed edges connecting each pair of feature nodes. 
To simplify the notation
$e_{u_iv_j}$, we use $e_{ij}$ in the context of feature matrix $\mathbf{D}$, and $e_{uv}$ in the context of graph $\mathcal{G}$.
In $\mathcal{E}_{OF}$,  
if $D_{ij}$ is a categorical variable, $e_{ij}$ is initialized with the one-hot vector corresponding to $D_{ij}$. On the other hand, if $D_{ij}$ is a continuous variable, $e_{ij}$ is initially taken as $D_{ij}$. In $\mathcal{E}_{FF}$, the directed edge $e_{v_i\rightarrow v_{j}}$ is regarded as the message passing from $v_i$ to $v_{j}$  as detailed in Section \ref{Complete}.

\textbf{Feature imputation as edge prediction} Based on the graph representation of the data matrix, the feature imputation problem  is naturally transformed into an edge prediction task on the  graph. Specifically, the prediction of the $j$-th feature of the $i$-th sample can be expressed as predicting $e_{ij}$ in  $\mathcal{E}_{OF}$. Edge prediction is conducted by minimizing the difference between $\widehat{e}_{ij}$ and $e_{ij}$. When $e_{ij}$ corresponds to a discrete feature, the difference 
is measured using cross-entropy. In the case of a continuous feature, the difference is measured using Mean Squared Error (MSE).

\textbf{Label prediction task in BCGNN} When facing label prediction task, we use 
$\widehat{\mathbf{D}}=(\widehat{\mathbf{D}}_{i\cdot})$ predicted by the upstream BCGNN as intermediate input. Then $\widehat{{Y}}_i$ is obtained by using a simple forward propagation through a trainable mapping: $\widehat{{Y}}_i = f(\widehat{\mathbf{D}}_{i\cdot})$, $\forall i \in \{1,\ldots,n\}$. The mapping $f(\cdot)$ and upstream BCGNN can be learned by minimizing the difference between $\widehat{{Y}}_i$ and ${Y}_i$. When ${Y}_i$ is a discrete label, 
we use cross-entropy to measure the difference between $\widehat{{Y}}_i$ and ${Y}_i$. When ${Y}_i$ is a continuous label, we use MSE to measure it.  

\subsection{Bipartite Graph: Data-Driven Inductive Learning}

\label{bipartite}

During the graph learning, information from data is exclusively conveyed by edges connecting feature nodes and observation nodes, which are established based on the observed data. Inspired by the GraphSAGE architecture \cite{hamilton2018inductive}, we propose to use the bipartite graph $\mathcal{G}_B=(\mathcal{V}, \mathcal{E}_{OF})$ \cite{you2020handling}. This framework shares similarities with GraphSAGE but distinguishes itself by the additional incorporation of edge embeddings. 
In $\mathcal{G}_B$, each node aggregates information from neighboring  nodes and those corresponding edge embeddings for inductive learning. 

Specifically, at each GNN layer $l$, to inductively learn the information from neighboring nodes in $\mathcal{G}_B$, observation and feature nodes aggregate messages from the embeddings of their neighboring nodes as well as the corresponding edge embeddings as follows: $\forall v\in\mathcal{V}$,
\begin{equation*}
	\small
	m_v^{l} = \mbox{AGG}_l\left\{ \begin{aligned}
		&\sigma(P^{l}\cdot \mbox{CONCAT}(h_v^{l-1},e_{uv}^{l-1},h_u^{l-1})+b^l)\big|\\
		&\forall u \in \mathcal{N}(v,\mathcal{G}_{B})
	\end{aligned} \right\}
\end{equation*}
with $h_u^{l-1} \in \mathbb{R}^{d_n}$,  $h_v^{l-1} \in \mathbb{R}^{d_n}$ and  $e_{uv}^{l-1} \in \mathbb{R}^{d_e}$, where    $P^l\in\mathbb{R}^{d_m\times{(2d_n+d_e)}}$ is a trainable weight matrix,  $b^l\in\mathbb{R}^{d_m}$ is a trainable bias term, $\sigma$ is an activation function, $\mbox{AGG}_l$ is the aggregation function, and $d_n$, $d_e$, and $d_m$ respectively denote the dimensions of the node, edge embedding and message. $\mathcal{N}(v,\mathcal{G})$ denotes the set of neighboring nodes of node $v$ in graph $\mathcal{G}$. 
Subsequently, the embeddings of observation and feature nodes are updated based on aggregated messages:
\begin{equation*}
	h_v^l=\sigma(Q^l\cdot{\mbox{CONCAT}(h_v^{l-1},m_v^{l}})+c^l), \forall v \in \mathcal{V},
\end{equation*}
where $Q^l\in\mathbb{R}^{d_n\times{(d_n+d_m)}}$ is a trainable weight matrix, and $c^l\in\mathbb{R}^{d_n}$ is a trainable bias term. After the node embeddings are updated, the edge embeddings will also be updated by 
\begin{equation*}
	e_{uv}^l=\sigma(W^l\cdot{\mbox{CONCAT}(e_{uv}^{l-1},h_v^{l},h_u^{l}})+d^l),\forall e_{uv}\in{\mathcal{E}_{OF}},
\end{equation*}
where $W^l\in\mathbb{R}^{d_e\times(2d_n+d_e)}$ is a trainable weight matrix, and $d^l\in\mathbb{R}^{d_e}$ is a trainable bias term.

\subsection{Complete Directed Graph: Feature Interdependence Structure Learning}
\label{Complete}
The interdependence structure among features holds valuable information for imputing missing features, yet this aspect is currently neglected by existing methods. The complete directed graph $\mathcal{G}_C=(\mathcal{V}_F, \mathcal{E}_{FF})$ proposed in this subsection is specifically constructed to uncover the interdependence structure among features. Specifically, when dealing with a target feature node $v$, all $w \in \mathcal{V}_F$ with $w\not=v$ are considered as source nodes under the perspective of message passing. Our approach involves learning the directed dependence structure between $h_v$ and each source feature node embedding $h_w$. This structure is characterized by message passing from $h_w$ to $h_v$, with the direction from $w$ to $v$. 

To measure the varying degree of dependency of the target node feature $v$ on different segments of the source node embedding $h_w$ in the complete directed graph $\mathcal{G}_C$, we employ an element-wise attention weight vector $\alpha_{w\rightarrow{v}}$. This weight vector is trainable and can be acquired through an additive attention mechanism. In detail, at each GNN layer $l$, we initially implement an additive attention mechanism $\mathcal{A}: \mathbb{R}^{2d_n} \rightarrow \mathbb{R}^{d_n}$ on $\mbox{CONCAT}(h_w^{l-1}, h_v^{l-1})$ to derive the attention score vector $\mbox{Score}_{w\rightarrow{v}}^{l}$, where $\mathcal{A}$ represents a single linear layer equipped with a trainable weight matrix $U_w^{l} \in \mathbb{R}^{d_n\times{2d_n}}$, followed by a LeakyReLU activation function. In other words, the vector that represents the attention weight score of the target feature $v$ on the source feature $w$ can be expressed as
\begin{equation*}
	\small
	\mbox{Score}^{l}_{w\rightarrow{v}} = \sigma({U^{l}_w} \cdot \mbox{CONCAT}({h^{l-1}_w},{h^{l-1}_v})+g^l), 
\end{equation*}
where $\sigma$ is the LeakyReLU activation function and $g^l\in\mathbb{R}^{d_n}$ is a trainable bias term. It is noted that, across different target node features $v$, the weight matrix $U_w$ remains shared for a given source node $w$. Subsequently, the attention score $\mbox{Score}_{w\rightarrow{v}}$ undergoes normalization through the softmax function to yield the element-wise attention weight vector $\alpha_{w\rightarrow{v}}$. Specifically, the $k$-th element of the normalized attention weight vector $\alpha_{w\rightarrow{v}}^l$ is written as
\begin{equation*}
	\small
	\alpha^{l}_{w\rightarrow{v}}[k]=\frac{\exp(\mbox{Score}_{w\rightarrow{v}}^l[k])}{\sum_{k'\in\{1,2,\ldots,d_n\}}\exp(\mbox{Score}^{l}_{w\rightarrow{v}}[k'])}.
\end{equation*}

Next, we calculate the Spearman correlation coefficient between features $w$ and $v$. Let $\rho_{wv}$ indicate the sign of the calculated Spearman correlation coefficient, which serves as an indicator of whether
the message passing from node $w$ to node $v$ is positive
or negative. The details of the calculation are deferred to Appendix \ref{Spearman}.
Furthermore, when considering a specific target feature node $v$ aggregating messages, the strength of messages from different source nodes should vary because the dependence relationships among features are distinct. We thus introduce a trainable scalar parameter $I_{w\rightarrow{v}}^l$ to quantify the strength of message passing from each source node $w$ to the target node $v$ in $\mathcal{G}_C$. 

Finally, taking into account the attention weight vector $\alpha_{w\rightarrow{v}}^l$, we represent the dependency of the target feature $v$ on the source feature $w$ as
\begin{equation*}
	\mbox{Message}^{l}_{w\rightarrow{v}} = (\rho_{wv}\cdot I^{l}_{w\rightarrow{v}}\cdot{\alpha^{l}_{w\rightarrow{v}}})\odot{h^{l-1}_w},
\end{equation*} 
where $\odot$ is the Hadamard product.
$\mbox{Message}^{l}_{w\rightarrow{v}}$ is the so-called message passing from $h_w$ to $h_v$ at  GNN layer $l$.
Exchanging unique messages bidirectionally between each pair of feature nodes effectively captures the directed dependency structure among features.

\subsection{Union Graph: Global Learning for Feature Imputation and Label Prediction} 
\label{union}

After the generation and transmission of messages in the bipartite graph $\mathcal{G}_B$ and the complete directed graph $\mathcal{G}_C$, we introduce the union graph. In the union graph, feature nodes aggregate messages from both neighboring observation and feature nodes, while observation nodes aggregate messages from neighboring feature nodes. 
Specifically, at each GNN layer $l$, we represent the global message aggregation process in BCGNN as follows:
\begin{equation*}
	\small
	m_v^{l} = \mbox{AGG}_l \left\{ \begin{aligned}
		&\sigma(P^{l}\cdot \mbox{CONCAT}(h_v^{l-1},e_{uv}^{l-1},h_u^{l-1})\\&~~+b^l), 
		\rho_{wv}\cdot I^l_{w\rightarrow{v}}\cdot\alpha^{l}_{w\rightarrow{v}}\odot{h^{l-1}_w}\big| \\
		&\forall u \in \mathcal{N}(v,\mathcal{G}_{B}),\forall w \in \mathcal{N}(v,\mathcal{G}_{C})
	\end{aligned} \right\}, \forall v\in\mathcal{V}_F,
\end{equation*}
\begin{equation*}
	\small
	m_v^{l} = \mbox{AGG}_l\left\{ \begin{aligned}
		&\sigma(P^{l}\cdot \mbox{CONCAT}(h_v^{l-1},e_{uv}^{l-1},h_u^{l-1})\\&~~+b^l)\big|
		\forall u \in \mathcal{N}(v,\mathcal{G}_{B})
	\end{aligned} \right\},\forall v\in\mathcal{V}_O.
\end{equation*}
After the message passing and aggregation, the embedding of every node $v \in \mathcal{V}$ is updated using $h^{l-1}_v$ and the aggregated message $m_v^{l}$ as follows:
\begin{equation*}
	\small
	h_v^l=\sigma(Q^l\cdot{\mbox{CONCAT}(h_v^{l-1},m_v^{l}})+c^l). 
\end{equation*}

Then, BCGNN updates the embedding of edges $e_{uv} \in \mathcal{E}_{OF}$:
\begin{equation*}
	\small
	e_{uv}^l=\sigma(W^l\cdot{\mbox{CONCAT}(e_{uv}^{l-1},h_v^{l},h_u^{l}})+d^l), 
\end{equation*}

which serves as the input for generating messages in the next layer of BCGNN. After BCGNN undergoes forward propagation through $L$ layers of updates, a simple linear readout layer is applied for the final feature imputation: 
\begin{equation*}
	\small
	\widehat{e}_{uv}=O_{\mbox{feature}}(h_{u}^{L},h_{v}^{L}).
\end{equation*}
This process yields $\widehat{\mathbf{D}}$. 
For a label prediction task related to the $i$-th observation, a linear readout layer is applied to the combined imputed feature edges of the $i$-th observation:
\begin{equation*}
	\widehat{{Y}}_{i}=O_{\mbox{label}}(\widehat{\mathbf{D}}_{i\cdot}).
\end{equation*}

We use distinct $d_n$-dimensional one-hot vectors to initialize the embedding $h_v$ for each feature node in $\mathcal{V}_F$. For observation nodes in $\mathcal{V}_O$, the embedding $h_u$ is initialized using the $d_{n}$-dimensional constant vector $\textbf{1}_{d_n}$.

\begin{figure}[t] 
	\centering 
	\includegraphics[height=4.3cm,width=12.9cm]{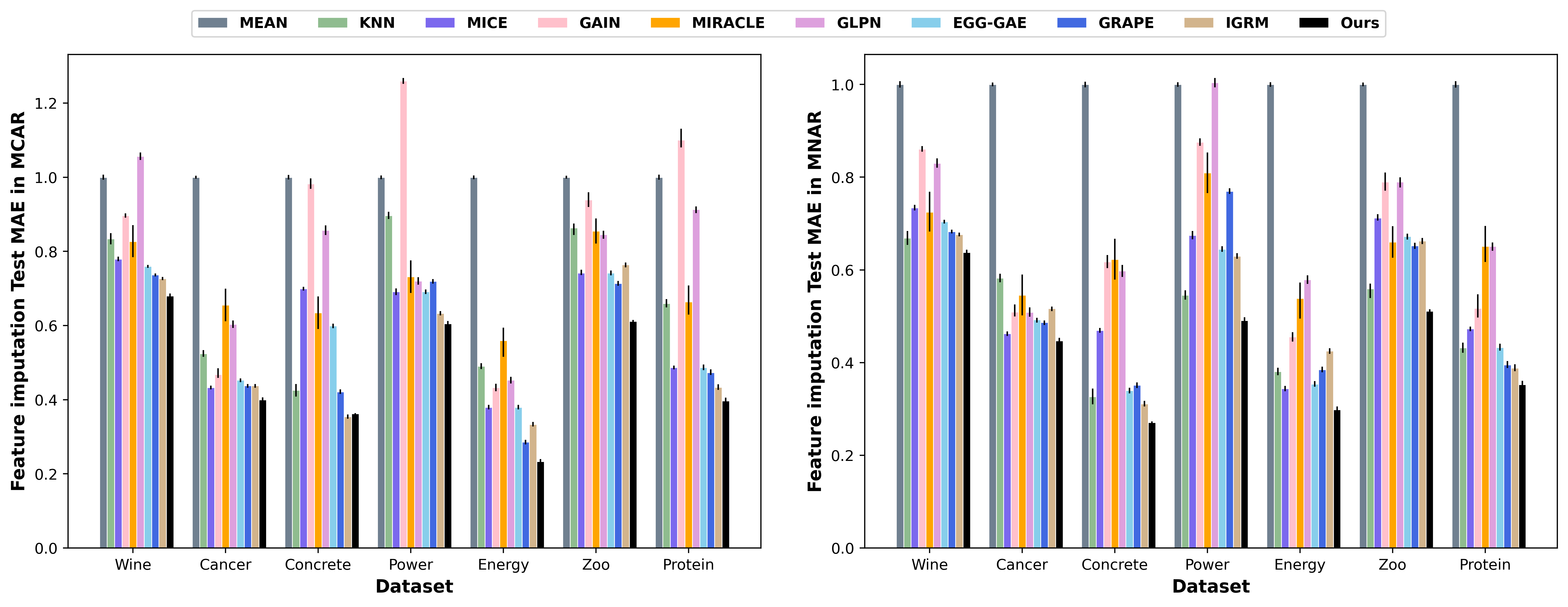}
	\caption{Average test MAE of feature imputation at a  missing rate of 0.3 under MCAR (Left) and MNAR (Right) in  UCI datasets over 5 random trials. The results are normalized by the average performance of the Mean imputation. 
	}
	\label{MAE}
\end{figure}
\setlength{\belowcaptionskip}{-5pt}

\subsection{Dropout Schemes in BCGNN}

To address the over-fitting problem in BCGNN, we use the DropEdge scheme \cite{rong2019dropedge} in the two subgraphs. In training phase, we randomly mask out some edges in both $\mathcal{E}_{OF}$ and $\mathcal{E}_{FF}$ before conducting forward propagation in each epoch:
\begin{equation*}
	\small
	\mbox{DropB}(\mathcal{E}_{OF},r_{B}) = \{e_{ij} | e_{ij} \in \mathcal{E}_{OF},  {M}^{B}_{ij} > r_{B},{M}_{ij}=1\}
\end{equation*}
\begin{equation*}
	\small
	\mbox{DropC}(\mathcal{E}_{FF},r_{C}) = \{e_{v_i\rightarrow{v_j}} | e_{v_i\rightarrow{v_j}} \in \mathcal{E}_{FF},  {M}^{C}_{ij} > r_{C}\}
\end{equation*}

where $ r_{B}$ and $ r_{C}$ are dropout rates, the mask matrices $\mathbf{M}^{B}=({M}^{B}_{ij}) \in \mathbb{R}^{n \times m} $ and $\mathbf{M}^{C}=({M}^{C}_{ij})\in\mathbb{R}^{c\times c}$ are both sampled uniformly within the interval $[0,1]$, with $c=m(m-1)$. Then, we only input those edges that are not masked into BCGNN for forward computation, and compute the training loss exclusively on the masked edges in $\mathcal{E}_{FF}$. This approach ensures that the predicted values of edges during training are definitely not functions of themselves, thus avoiding the problem of over-fitting. In the testing phase, the entire graph $\mathcal{G}$ without Dropedge is fed into BCGNN to conduct feature imputation. 
Furthermore, aim at improving the generalization capacity of BCGNN, we
incorporate the AttentionDrop technique in element-wise attention mechanism between feature nodes in $\mathcal{G}_{C}$: 
\begin{equation*}
	\mbox{AttentionDrop}(\alpha_{w\rightarrow{v}},a_{drop})  = \{\alpha_{w\rightarrow{v}}[k]\cdot I(b_k>a_{drop})\}, 
\end{equation*} 

where $b_k$  is uniformly sampled within $[0,1]$ and $a_{drop}$ is the dropout rate. The detailed forward computation of BCGNN 
can be found in Algorithm \ref{algorithm}.

\begin{algorithm}[t]
	\small
	\caption{BCGNN forward computation}
	\begin{algorithmic}
		\STATE {\bfseries Input:} Graph $\mathcal{G} = (\mathcal{V},\mathcal{E})$ with $\mathcal{V}=\mathcal{V}_F\cup\mathcal{V}_O$ and $\mathcal{E}=\mathcal{E}_{OF}\cup \mathcal{E}_{FF}$; Number of layers ${L}$; DropEdge rate $r_{B}$ and $r_{C}$; AttentionDrop rate $a_{drop}$; $\mathcal{N}(v,\mathcal{G})$ is a mapping providing the set of neighboring nodes of node $v$ within the graph $\mathcal{G}$; $P^{l}$ for message passing; $Q^{l}$ for node updating; $W^{l}$ for edge updating; ${U}_{w}^l$ for attention weight score calculation; $I_{w\rightarrow{v}}^l$ quantifying the strength of message passing from  node $w$ to node $v$.
		\FOR{$l$ {\bfseries in} $\{1,\ldots,L\}$}
		\STATE  $h^{(0)}_v \leftarrow \mbox{INIT}(v)$,$~~\forall{v\in\mathcal{V}}$ ; $e^{(0)}_{uv} \leftarrow \mbox{INIT}(e_{uv}),~~\forall{e_{uv}\in{\mathcal{E}}}$ 
		\STATE $\mathcal{E}_{{dropB}},\mathcal{E}_{{dropC}}$  $\leftarrow$ $\mbox{DropB}(\mathcal{E}_{OF},r_{B})$, $\mbox{DropC}(\mathcal{E}_{FF},r_{C})$; $\mathcal{G}_{dropB}$, $\mathcal{G}_{dropC}$ $\leftarrow$ $(\mathcal{V},\mathcal{E}_{dropB})$, $(\mathcal{V}_F,\mathcal{E}_{dropC})$;
		\FOR{$v$ $\in$ $\mathcal{V}$}
		
		\FOR{$u$ $\in$ $\mathcal{N}(v,\mathcal{G}_{dropB})$}
		\STATE $\mbox{MessageB}^{l}_{u{v}}= \sigma(P^{l}\cdot{\mbox{CONCAT}(h_v^{l-1},e_{uv}^{l-1},h_u^{l-1})}+b^l)$
		\ENDFOR
		\IF{$v \in \mathcal{V}_{F}$}
		\FOR{$w$ $\in$ $\mathcal{N}(v,\mathcal{G}_{dropC})$}
		\STATE $\mbox{Score}_{w\rightarrow{v}}^{l} = \sigma({U_w^{l}}\cdot\mbox{CONCAT}(h_w^{l-1},h_v^{l-1})+g^l)$
		\STATE $\alpha^{l}_{w\rightarrow{v}}= \mbox{AttentionDrop}( \mbox{Softmax}(\mbox{Score}_{w\rightarrow{v}}^{l}),a_{drop})$
		\STATE $\mbox{MessageC}^{l}_{w\rightarrow{v}} = (\rho_{wv}\cdot I_{w\rightarrow{v}}^l\cdot\alpha^{l}_{w\rightarrow{v}})\odot{{h^{l-1}_w}}$ 
		\ENDFOR
		\ENDIF
		\IF{$v \in \mathcal{V}_{O}$}
		\STATE $m_v^{l}=\mbox{AGG}_l(\mbox{MessageB}^{l}_{uv}|\forall{u}\in{\mathcal{N}(v,\mathcal{G}_{dropB})})$
		\ELSIF{$v \in \mathcal{V}_{F}$}
		\STATE 	$m_v^{l}=\mbox{AGG}_l(\mbox{MessageB}^{l}_{uv}, \mbox{MessageC}^{l}_{w\rightarrow{v}}|\forall{u}\in{\mathcal{N}(v,\mathcal{G}_{dropB}),\forall{w\in{\mathcal{N}(v,\mathcal{G}_{dropC})}}})$
		\ENDIF
		\STATE
		$h_v^{l}=\sigma(Q^{l}\cdot{\mbox{CONCAT}(h_v^{l-1},m_v^{l})}+c^l)$
		\ENDFOR
		\FOR{$e_{uv}$ $\in$ $\mathcal{E}_{dropB}$}
		\STATE $e^{(l)}_{uv}= \sigma(W^l\cdot{\mbox{CONCAT}(h_v^{l},e_{uv}^{l-1},h_u^{l})}+d^l)$
		\ENDFOR
		\ENDFOR
		\STATE {\bfseries Output for Feature Imputation}: $\widehat{e}_{uv}=O_{\mbox{feature}}(h_{u}^{L},h_{v}^{L})$, $\forall{e_{uv}\in{\mathcal{E}_{OF}}}$
		\STATE {\bfseries Output for Label Prediction}: $\widehat{{Y}} _{i}=O_{\mbox{label}}(\widehat{\mathbf{D}}_{i\cdot}),\forall{i}\in\{1,\ldots,n\}$
		
	\end{algorithmic}
	\label{algorithm}
\end{algorithm}

\section{Experiments}
\label{experiment}
\subsection{Datasets and Baseline Models}

\textbf{Datasets.} We conduct experiments on seven datasets from the UCI Machine Learning Repository. These datasets span diverse domains including medicine (Cancer), civil engineering (Concrete, Energy), industry (Power), business (Wine), and biology (Protein, Zoo). Cancer has the highest feature dimension, 
consisting of 32 features
with 569 observations. Protein has the largest sample size, which contains over 45,000 observations. The Wine, Concrete, Protein, and Energy datasets include a mix of discrete and continuous features. Zoo exclusively comprises discrete features, while the Power and Cancer datasets only have continuous features. The feature values in all dataset are scaled to $[0, 1]$ with a MinMax scaler \cite{leskovec2020mining}.

\textbf{Configurations.} 
Since the datasets are  fully observed,  we simulate missing data by generating a masking matrix $\mathbf{M}$ according to each distinct missing mechanism and deleting observed values from the data matrix based on $\mathbf{M}$. We consider three types of missing mechanisms: missing completely at random (MCAR), missing at random (MAR), and missing not at random (MNAR) 
\cite{rubin1976inference}. The detailed procedures for generating the masking matrices for various missing mechanisms are provided in Appendix \ref{generatemiss}, and
the configurations of BCGNN can be found in Appendix \ref{configuration}.

\begin{figure*}[t]
	\centering
	\begin{minipage}{0.50\textwidth}
		\centering
		\includegraphics[height=3cm,width=8.2cm]{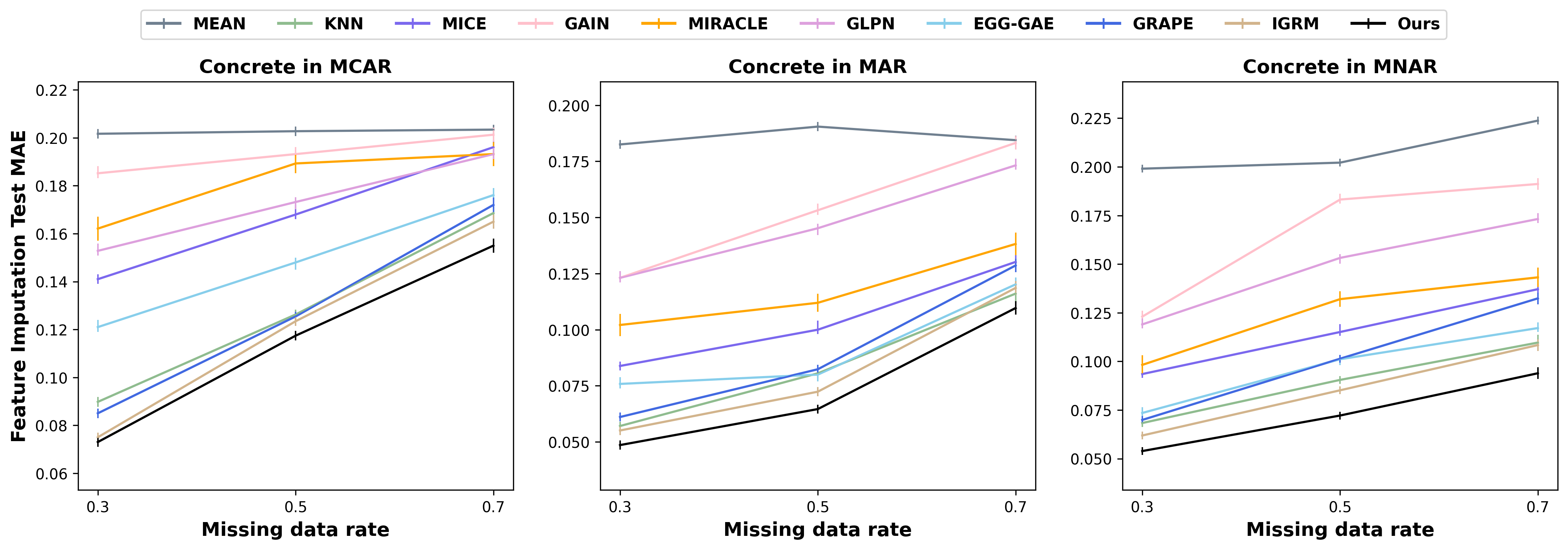}
	\end{minipage}%
	\hfill
	\hfill
	\begin{minipage}{0.38\textwidth}
		\centering
		\includegraphics[height=2.8cm,width=4.8cm]{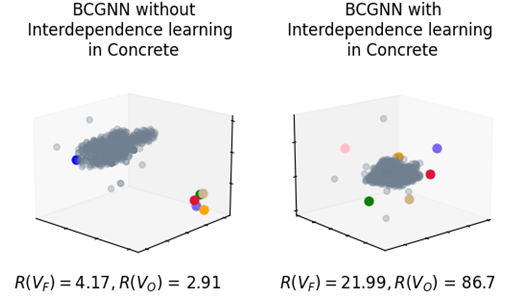}
	\end{minipage}
	\begin{minipage}{0.50\textwidth}
		\centering
		\includegraphics[height=3cm,width=8.2cm]{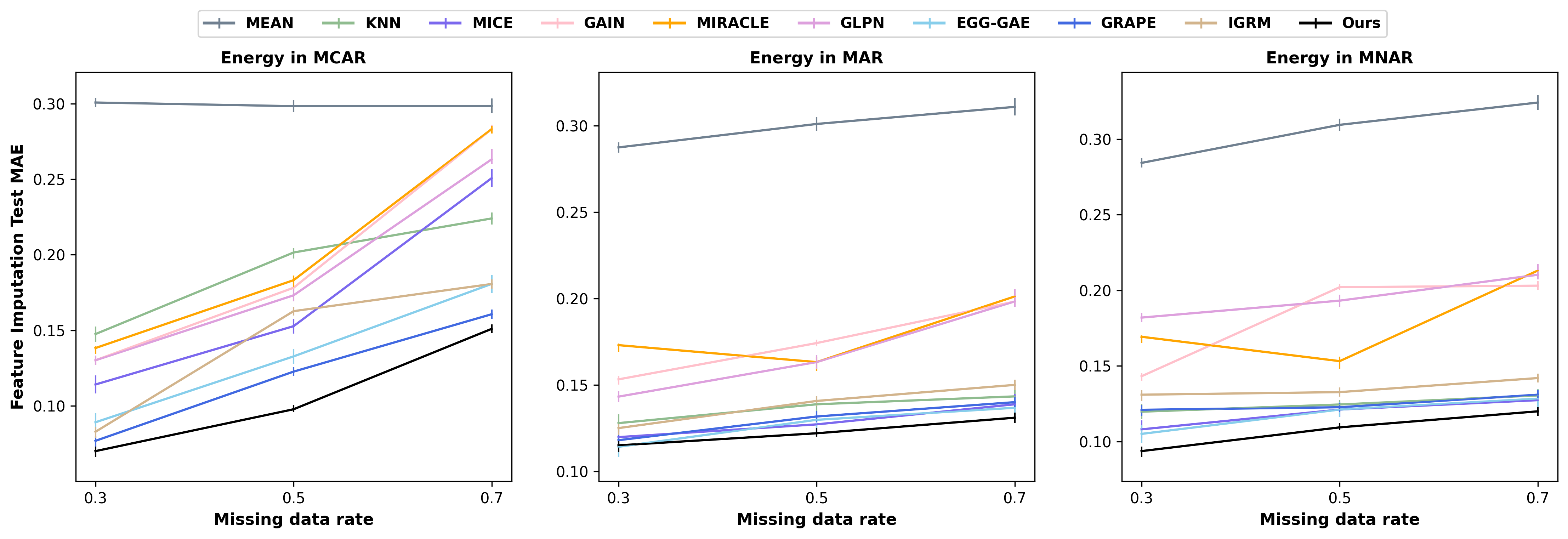}
	\end{minipage}%
	\hfill
	\hfill
	\begin{minipage}{0.38\textwidth}
		\centering
		\includegraphics[height=2.8cm,width=4.8cm]{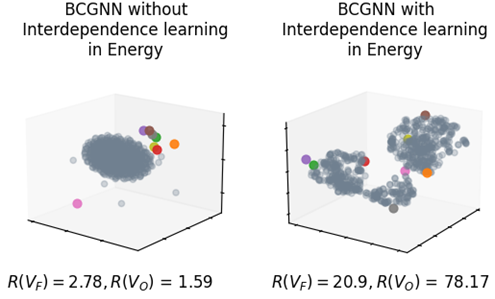}
	\end{minipage}
	
	\caption{\textbf{Left}: Average test MAE of feature imputation under MCAR, MAR and MNAR with different missing rates in the Concrete and Energy datasets over 5 random trials. \textbf{Right}: The embedding spaces $V_F$ and $V_O$ for feature and observation nodes, respectively. They are obtained from the trained BCGNN with/without learning interdependence structure in Concrete and Energy under MAR with a missing rate of 0.3. The \textbf{colored} dots represent feature node embeddings and the \textbf{grey} dots represent observation node embeddings.}
	\label{A}
\end{figure*}

\textbf{Baseline models.} In our experiments, we compare BCGNN with 10 baseline models: (1) Mean: imputation using
the feature-wise mean; (2) KNN \cite{kim2004reuse}; (3) MICE \cite{van2011mice}; (4) GAIN \cite{yoon2018gain}; (5) Decision Tree: we only use it in the label prediction task \cite{loh2011classification}; (6) MIRACLE: we only use it in the feature imputation task \cite{kyono2021miracle}; (7) GRAPE \cite{you2020handling}; (8) GLPN: a method has demonstrated  superiority over VGAE and GDN \cite{zhang2023missing}, and is only used in the feature imputation task; (9) EGG-GAE \cite{telyatnikov2023egg}; (10) IGRM \cite{zhong2023data}.

\subsection{Feature Imputation}\label{featureimputation}
At a missing rate of 0.3, we compare the feature imputation performance of BCGNN with the nine baseline models on the seven datasets under the three missing mechanisms. We record the averaged test error for each method over five random trials. The test error is defined as the mean absolute error (MAE) between $D_{ij}$ and $\widehat{D}_{ij}$ over all entries with ${M}_{ij}=0$. The results under MCAR and MNAR are presented in Figure \ref{MAE}, and the results for MAR can be found in Figure \ref{featureimputationMAR} in Appendix \ref{comprehensive}. Compared with the best baseline model GRAPE or IGRM, our BCGNN  achieves an average reduction of \textbf{11.8\%}, \textbf{13.2\%}, and \textbf{16.1\%} in the test MAE for MCAR, MAR and MNAR, respectively.

We further consider  missing rates of 0.5 and 0.7. At a missing rate of 0.5, compared with the best baseline, GRAPE or IGRM, BCGNN reduces the  test MAE by an average of $13.5\%$, $12.3\%$, and $16.1\%$ under MCAR, MAR and MNAR respectively. At a missing rate of 0.7, it reduces the test MAE by $10.2\%$,  $11.4\%$ and $12.6\%$, respectively.
Figure \ref{A} (left panel) shows the specific results for feature imputation under various missing rates and missing mechanisms in Concrete and Energy datasets, and the results for all the other datasets are available in Figure \ref{featureimputationgraph} in Appendix \ref{comprehensive}.

\subsection{Feature Embedding Ability of BCGNN}

Accordin to \citet{jegelka2022theory} and \citet{zhang2023expressive}, the feature embedding ability of a GNN can be expressed by the size of the embedding space $V$, which is the embedding output space obtained from the GNN mapping. Here we use lossy coding to measure the size of $V$ with ${V}= (h_{v_1}, h_{v_2}, \ldots, h_{v_m}) \in \mathbb{R}^{m \times d_n}$. The lossy coding scheme maps $V$ to a sequence of binary bits, such that the original embeddings can be recovered up to an allowable distortion: $E[\|h_{v_i}-\widehat{h}_{v_i}\|^2] \leq \epsilon^2$. Then the size of $V$ can be quantified as: 
$R(V)=0.5\log_2\det\left(I+mVV^T/(\epsilon^2d_n)\right)$\cite{ma2007segmentation}.

We obtain the output embeddings from the experiments conducted in Section \ref{featureimputation} for the regular BCGNN and the BCGNN without interdependence structure learning. The latter is achieved by setting all $\rho_{wv}$ between feature nodes to zero, effectively disabling the message passing between feature nodes. Then we evaluate the feature embedding ability of the two types of BCGNN by comparing the sizes of their feature node embedding space $R(V_F)$ and observation node embedding space $R(V_O)$. We report the results for all datasets under different missing mechanisms in Figure~\ref{A} (right panel), Figures \ref{featureimputationgraph} (right panel),  \ref{TSNEMCAR} and \ref{TSNEMNAR} in the Appendix.  Clearly, the BCGNN with learned interdependence structure among features exhibits larger feature and observation embedding spaces than the BCGNN without it. This directly indicates that the learning of interdependence structure indeed improves the feature embedding ability of BCGNN.

\subsection{Generalization on New Observations}
When addressing  the feature imputation task for new observations, we hope to directly use their observed features  as the input to impute  missing values,  without the need for retraining the model. 
Thus we  investigate  the generalization  ability  of BCGNN in handling new  observations. In our experiments, we randomly divide the entire dataset $\mathbf{D} \in \mathbb{R}^{n \times m}$ into two groups of equal sample sizes: $\mathbf{D}^{old} \in \mathbb{R}^{n_{old} \times m}$ and $\mathbf{D}^{new} \in \mathbb{R}^{n_{new} \times m}$, representing the initial and new observations, respectively.
As detailed in Appendix \ref{generatemiss}, we respectively generate masking matrices $\mathbf{M}^{old}$ and $\mathbf{M}^{new}$ for $\mathbf{D}^{old}$ and $\mathbf{D}^{new}$ under the three missing mechanisms with a missing rate of 0.3. Then we train BCGNN on the initial observations $D^{old}_{ij}$ with $M^{old}_{ij}=1$. During the test phase, we employ the new observations $D_{ij}^{new}$ with $M^{new}_{ij}=1$ as the input for the trained BCGNN and directly impute ${\widehat{D}}^{new}_{ij}$ with $M^{new}_{ij}=0$. We repeat the same process for the baseline models requiring training (MIRACLE, GAIN, GRAPE, IGRM, EGG-GAE and GLPN). For the other baselines without training, we directly employ them on $\mathbf{D}^{new}$ to impute ${\widehat{D}}^{new}_{ij}$ with $M^{new}_{ij}=0$. As shown in Figure \ref{generalization} in Appendix \ref{labelpredictionandgeneralization}, BCGNN demonstrates outstanding generalization ability on new observations, which yields \textbf{11.3\%}, \textbf{12.8\%}, and \textbf{14.2\%} lower MAE compared with the best baseline, GRAPE, under MCAR, MAR, and MNAR, respectively. 

\subsection{Label Prediction}
To evaluate the performance of BCGNN in label prediction task, we randomly split the target label $\mathbf{Y}$ into a training set $\mathbf{Y}_{train}$ and a test set $\mathbf{Y}_{test}$ with a sample size ratio of 7:3, which are respectively used as observed and missing dataset. 
BCGNN can then be trained by minimizing the difference between $\widehat{{Y}}_i$ and ${Y}_i$ over all $i$ with ${M^y_{i}=1}$. For all baseline models except the decision tree, since there are no corresponding end-to-end prediction measures, we first impute and obtain the completed data $\widehat{\mathbf{D}}$, 
and then use a fully connected layer on each $\widehat{\mathbf{D}}_{i\cdot}$ to predict $\widehat{{Y}}_i$. As shown in  Figure~\ref{Labelprediction} in Appendix \ref{labelpredictionandgeneralization}, BCGNN yields \textbf{8.8\%}, \textbf{6.8\%}, and \textbf{7.3\%} lower MAE compared with the best baseline, GRAPE, under MCAR, MAR, and MNAR respectively over the 7 datasets.

\subsection{DropEdge and AttentionDrop}

We evaluate the performance of BCGNN under three missing mechanisms with a missing rate of 0.3, which utilizes DropEdge on both bipartite graph and complete directed graph, as well as AttentionDrop on the complete directed graph. We evaluate the BCGNN variants without one, two, or three of these schemes. We report the averaged MAE for feature imputation under different scenarios in Table \ref{B} in Appendix \ref{Dropedge}. It shows that using DropEdge on the bipartite graph significantly improves BCGNN's performance by consistently reducing the average test MAE by 35\%, employing DropEdge on the complete directed graph also leads to a 5\% reduction in the average test MAE, while AttentionDrop on the complete directed graph achieves an average reduction of 3\% in the test MAE.

\section{Conclusion}
In this paper, we propose BCGNN, a union of bipartite graph and complete directed graph, which addresses feature imputation and label prediction tasks by inductively learning node representations and explicitly learning interdependence structure among features. The directed complete graph is designed to learn and express the interdependence structure, thereby enhancing our model's expressive capability. Compared to SOTA methods, our model demonstrates significant improvements in accuracy, and robustly adapts to various missing ratios, missing mechanisms, and unseen data points. 

\backmatter

\newpage
\begin{appendices}

\section{Spearman Correlation Coefficient Estimator and Its Sign Indicator}
\label{Spearman}
A positive  correlation coefficient between two features indicates a consistent relationship, where both  features tend to increase or decrease simultaneously. Conversely, a negative coefficient signifies an inverse relationship, meaning that when one  feature increases, the other tends to decrease. Therefore, in BCGNN, it is crucial to estimate the sign of the correlation between each pair of features to facilitate message passing among feature nodes. We opt for using Spearman correlation coefficient to measure this correlation.

The original formula for  estimating the  Spearman correlation coefficient between two features $X$ and $Y$ based on $n$ observations is:
\vspace{-1pt}
\begin{equation*} 
	\small 
	\centering 
	S_{xy} = \frac{\frac{1}{n} \sum_{i=1}^{n} (R(x_i) - \bar{R}(x)) \cdot (R(y_i) - \bar{R}(y))}{\sqrt{\frac{1}{n} \sum_{i=1}^{n} (R(x_i) - \bar{R}(x))^2 \cdot \frac{1}{n} \sum_{i=1}^{n} (R(y_i) - \bar{R}(y))^2}} 
\end{equation*}
where $R(x_i)$ represents the rank of $x_i$ among the $n$ observations for feature $X$. 
In practice, BCGNN employs a simplified equivalent formula for estimating the Spearman correlation:
\vspace{-5pt}
\begin{equation*} 
	\small 
	\centering 
	S_{xy} = 1 - \frac{6 \sum d_i^2}{n(n^2 - 1)}, 
\end{equation*}
where $d_i=R(x_i)-R(y_i)$.
When estimating the Spearman correlation coefficient between features $X$ and $Y$ the presence of  missing data, we consider only the observations where both $X$ and $Y$ are simultaneously recorded. In scenarios with high missing rates or complex missing mechanisms, the absolute value of the estimated Spearman correlation coefficient often fails to fully and accurately represent the strength of dependency between the two features \cite{crysdian2022behaviour}. Nevertheless, we  extract the signs of 
estimated Spearman correlation coefficients to facilitate message passing among feature nodes in BCGNN. Specifically, 
for  features $w$ and $v$ with an estimated Spearman correlation coefficient $S_{wv}$, we define the sign indicator $\rho_{wv}$ as follows: when $|S_{wv}|< 0.1$, $\rho_{wv}=0$;  when $|S_{wv}|\geq0.1$, $\rho_{wv}=1$ if $S_{wv}>0$, and $\rho_{wv}=-1$ if $S_{wv}<0$.

The selection of 0.1 as a threshold is based on a common consensus in numerous studies. It is generally accepted that when the absolute value of the Spearman correlation coefficient falls below 0.1, the correlation between the random variables is considered negligible. \cite{mukaka2012guide, schober2018correlation}. Moreover, when an estimated Spearman correlation coefficient computed from incomplete data is near zero, there is a realistic possibility that
its true counterpart has an opposite sign. Thus, we conservatively set its corresponding  sign indicator to 0, effectively circumventing potential inaccuracy in the sign estimation under high missingness and complex missing mechanisms. In this setting, only strong correlations are used to facilitate the feature imputation.

\section{Generating missingness}
\label{generatemiss}

Consider the $k$-th subject and the $i$-th feature. 
We assume that $M_{ki}$ follows  a Bernoulli distribution with success probability  $\pi_{ki}$. 
The following text describes our approach for creating synthetic datasets that adhere to MCAR, MAR and MNAR
patterns of missing data.

\subsection{Missing Completely At Random (MCAR)}

The missing probability $\pi_{ki}$ remains constant, and is set to the current experiment's missing rate.


\subsection{Missing At Random (MAR)}

We assume that  $\pi_{ki}$ is determined by the observed values of the previous $i-1$ features of the $k$-th subject (if observed). Specifically, 
\begin{equation*} 
	\small 
	\centering 
	\pi_{ki} = \frac{p(i) \cdot n \cdot \exp{(\sum_{j<i}w_{j}m_j D_{kj} + b_j(1-m_{j}))}}{\sum_{l=1}^n \exp(\sum_{j<i}w_{j}m_j D_{lj} + b_j(1-m_j))}, 
\end{equation*}
where $p(i)$ is the average missing rate of the $i$-th feature,  $w_j$ and $b_j$ are sampled from a uniform distribution $U(0,1)$ once for each dataset, and  $m_j$ represents the dependency of the missingness of $D_{ki}$ on $D_{kj}$.

\subsection{Missing Not At Random (MNAR)}

The missingness of $D_{ki}$ depends on its own value. Specifically, 
\begin{equation*} 
	\small
	\centering 
	\pi_{ki} = \frac{p(i) \cdot n \cdot \exp(-w_i D_{ki})}{\sum_{l=1}^n \exp(-w_i D_{li})},
\end{equation*}
where $p(i)$ is the average missing rate of the $i$-th feature, and $w_i$ is sampled from a uniform distribution $U(0,1)$, representing the dependency of the missingness of 
$D_{ki}$ on its own value. 

\section{Configurations of BCGNN} 
\label{configuration}
In the experiments across all datasets, we train BCGNN for 20,000 epochs using Adam optimizer with a learning rate of 0.001. The experiments follow the  parameter settings in 
\cite{you2020design} and \cite{zhou2019graph}, and use a three-layer BCGNN with both node embeddings and edge embeddings of 64 dimensions. The $\mbox{AGG}_l$ is implemented using a mean pooling function MEAN($\cdot$). ReLU is used as the activation function in message generation, node updates and edge updates, while LeakyReLU is used in additive attention. The DropEdge ratio is set to 0.5, and the AttentionDrop ratio is set to 0.3. $O_{\mbox{feature}}$ and $O_{\mbox{label}}$ are both implemented with a single linear layer. For all experiments, we run 5 trials with different random seeds and report the mean and standard deviation of the results.

\begin{figure}[h]
	\centering
	\begin{minipage}{1\textwidth}
		\centering
		\includegraphics[height=5cm,width=11cm]{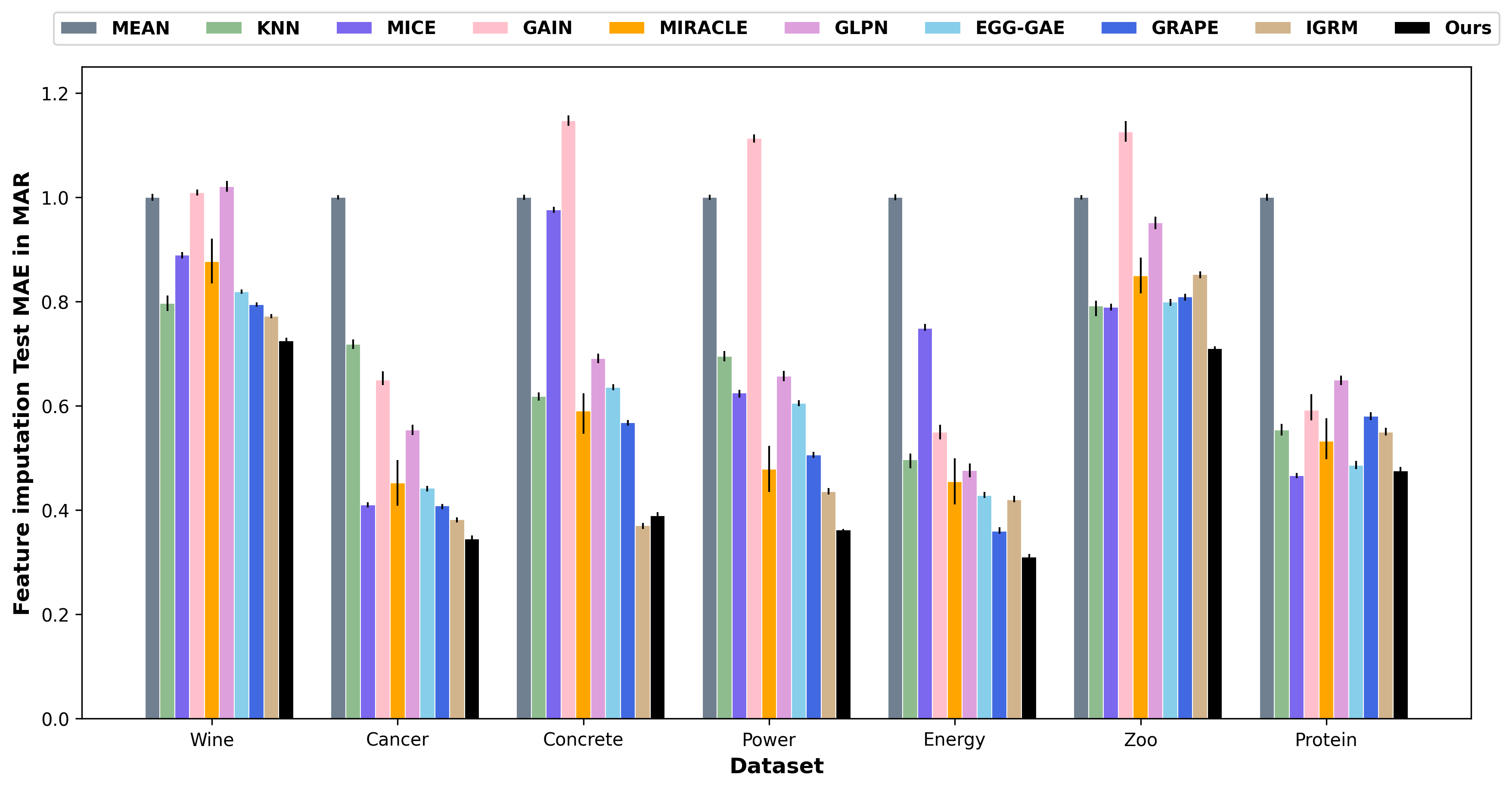}
	\end{minipage}%
	\caption{Average test MAE of feature imputation at a  missing rate of 0.3 under MAR in UCI datasets over 5 random trials. The results  are normalized by the average performance of the Mean imputation.}
	\label{featureimputationMAR}
	\vspace{-4mm}
	
\end{figure}

\begin{figure*}[t]
	\centering
	\begin{minipage}{0.50\textwidth}
		\centering
		\includegraphics[height=3cm,width=8cm]{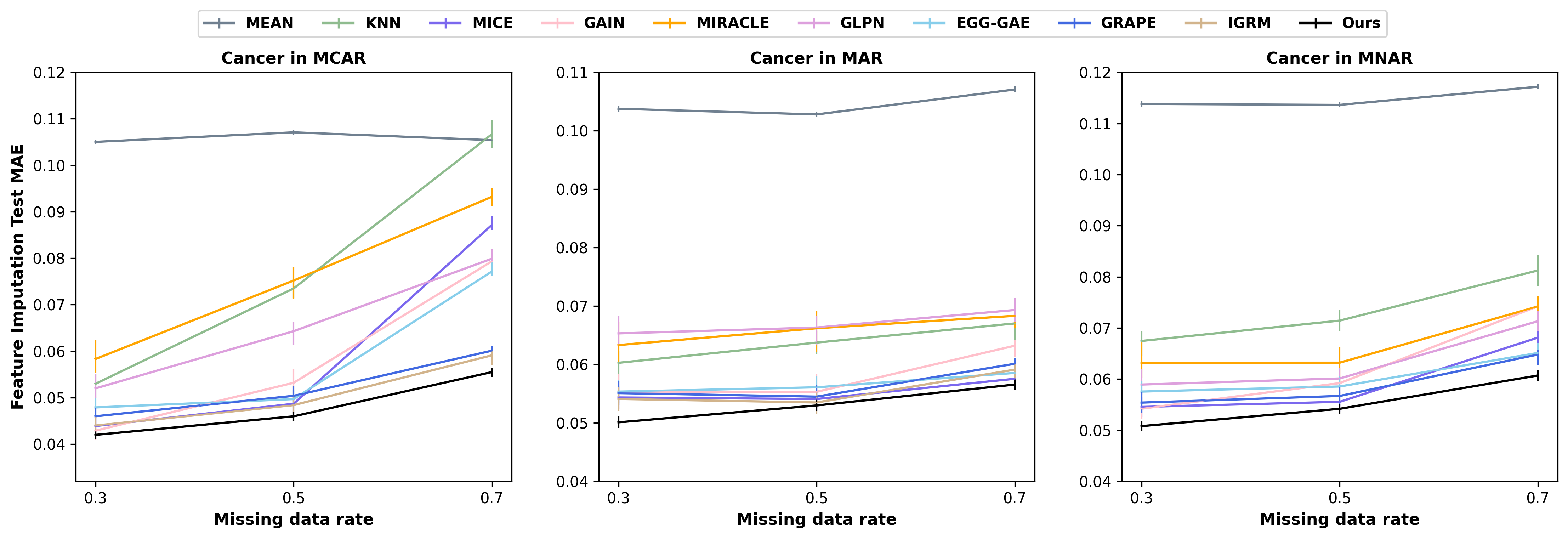}
	\end{minipage}%
	\hfill
	\begin{minipage}{0.43\textwidth}
		\centering
		\includegraphics[height=2.7cm,width=4.5cm]{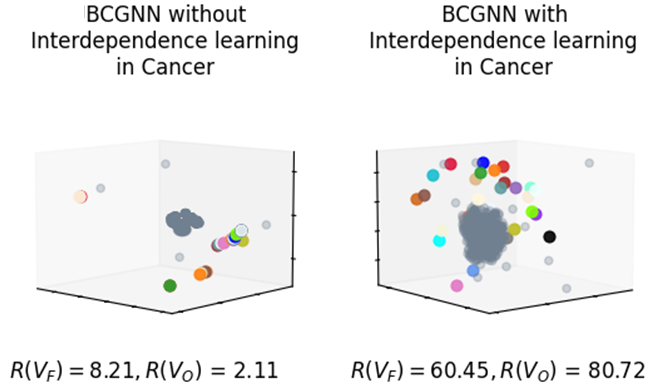}
	\end{minipage}
	\vspace{3pt} 
	\begin{minipage}{0.50\textwidth}
		\centering
		\includegraphics[height=3cm,width=8cm]{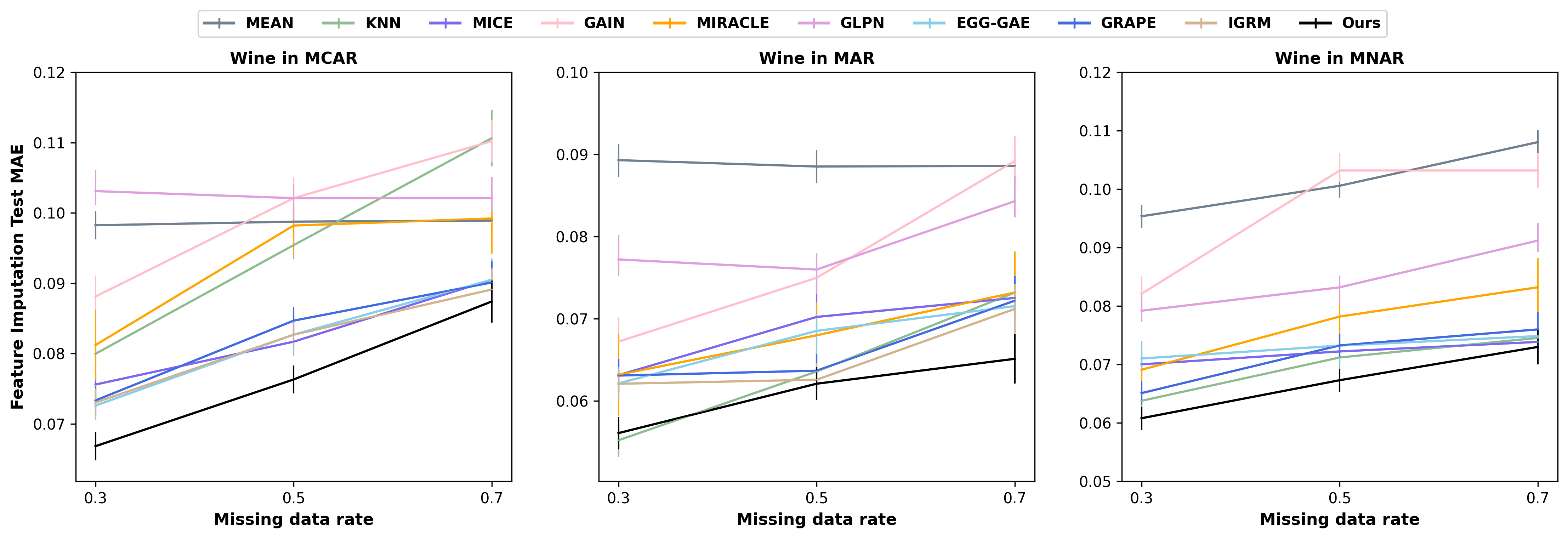}
	\end{minipage}
	\hfill
	\begin{minipage}{0.43\textwidth}
		\centering
		\includegraphics[height=2.7cm,width=4.6cm]{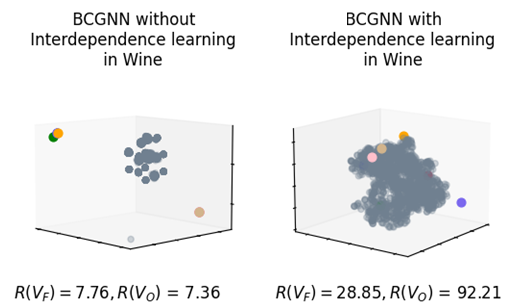}
	\end{minipage}
	\vspace{3pt}
	\begin{minipage}{0.50\textwidth}
		\centering
		\includegraphics[height=3cm,width=8cm]{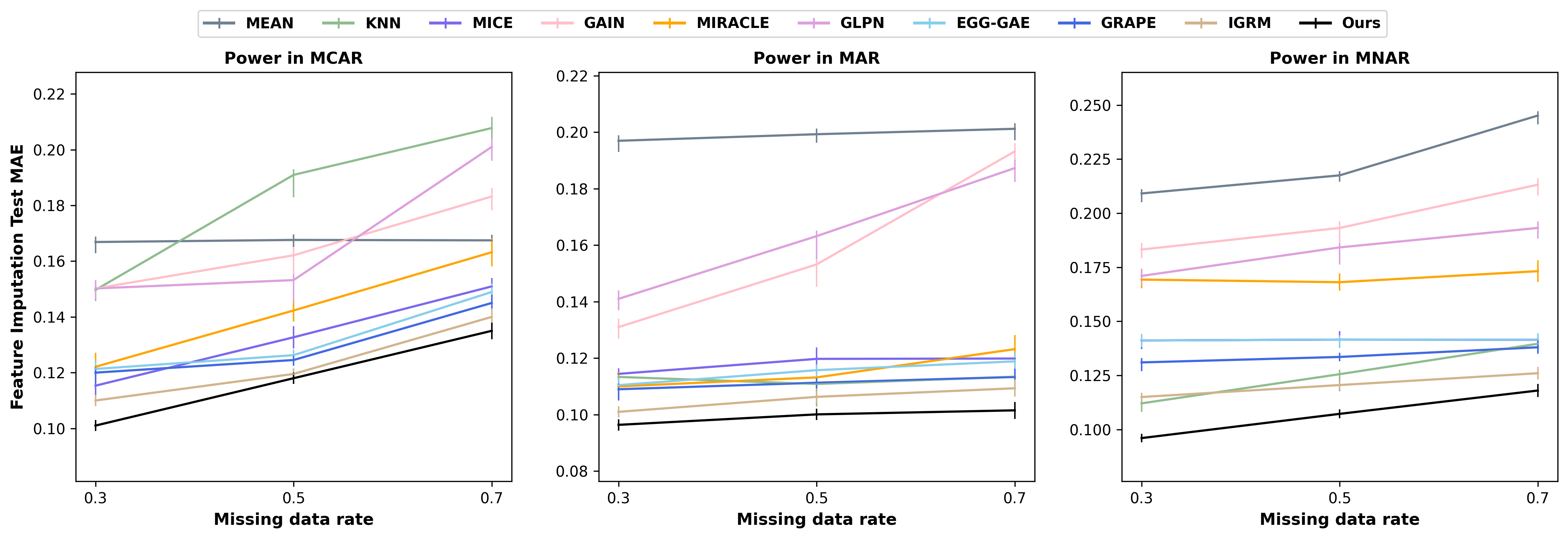}
	\end{minipage}
	\hfill
	\begin{minipage}{0.43\textwidth}
		\centering
		\includegraphics[height=2.7cm,width=4.6cm]{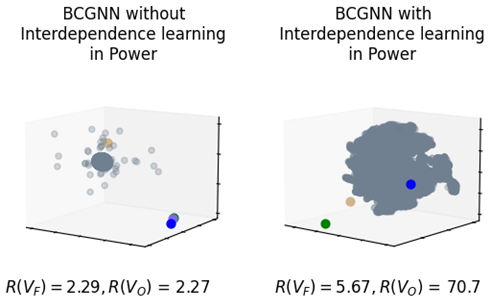}
	\end{minipage}
	\vspace{3pt}
	\begin{minipage}{0.50\textwidth}
		\centering
		\includegraphics[height=3cm,width=8cm]{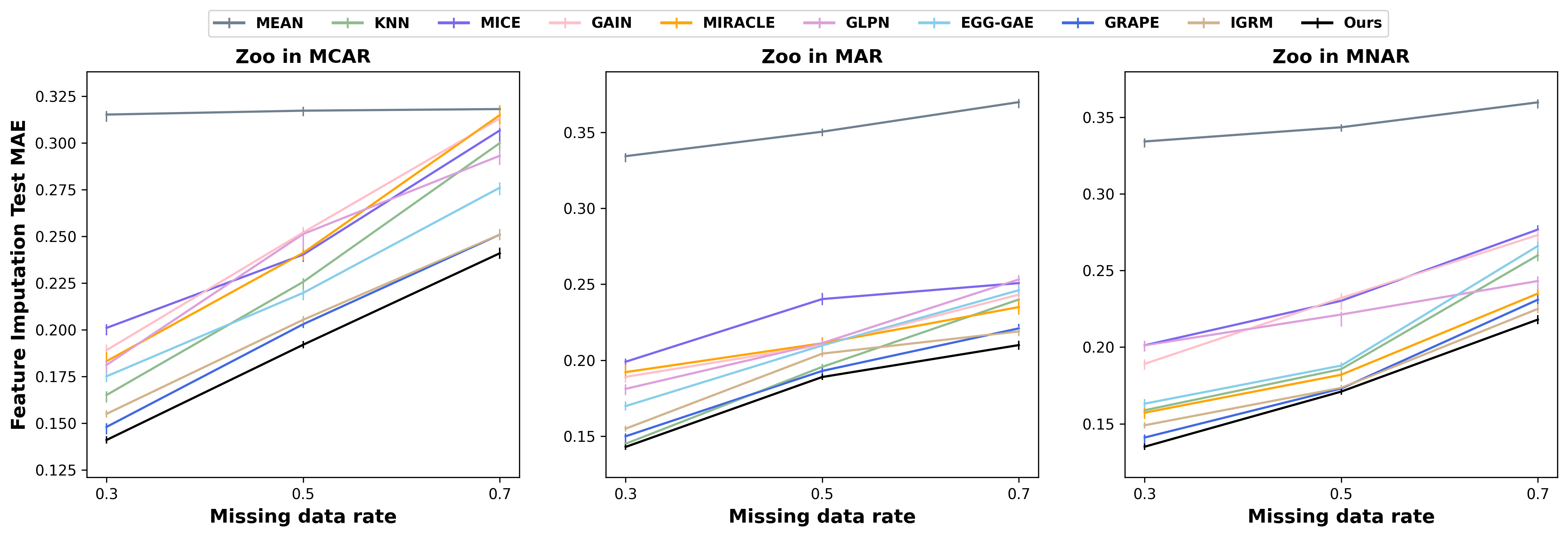}
	\end{minipage}
	\hfill
	\begin{minipage}{0.43\textwidth}
		\centering
		\includegraphics[height=2.7cm,width=4.6cm]{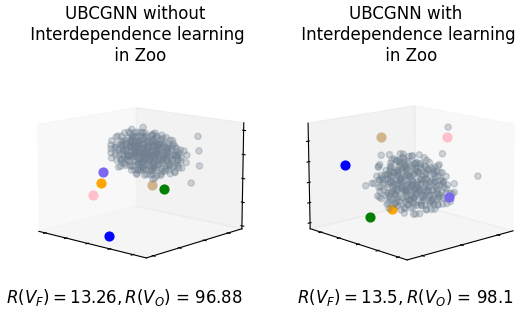}
	\end{minipage}
	\vspace{3pt}
	\begin{minipage}{0.50\textwidth}
		\centering
		\includegraphics[height=3cm,width=8cm]{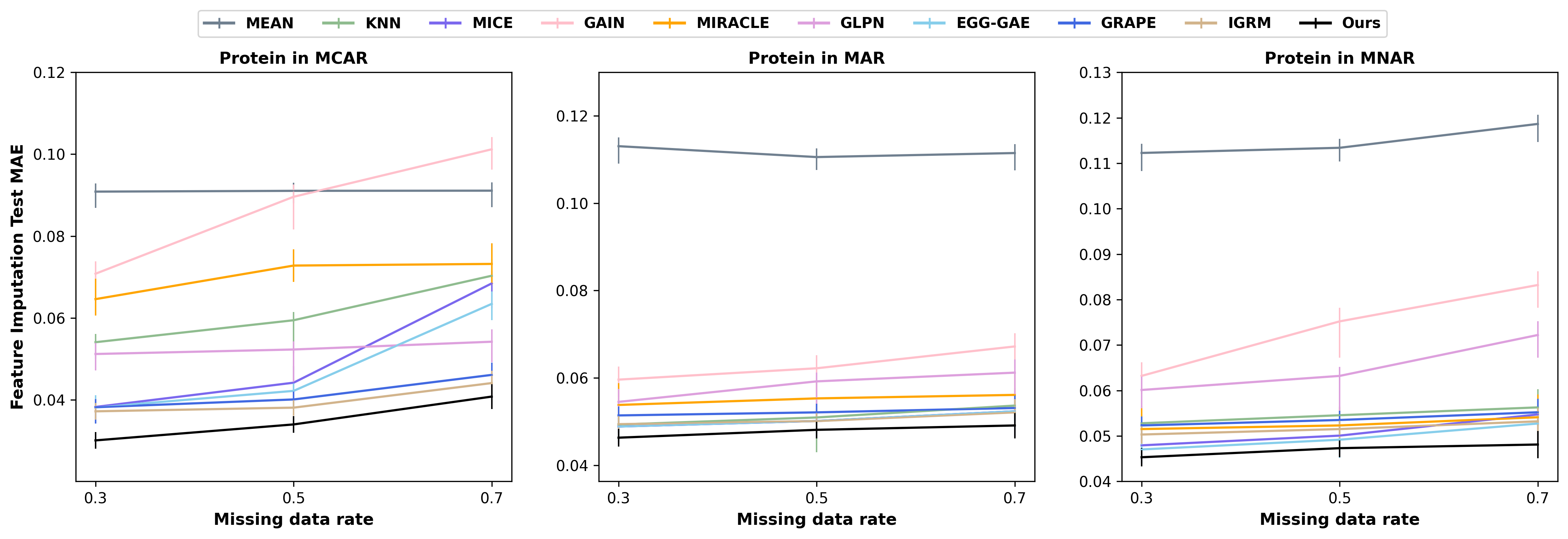}
	\end{minipage}
	\hfill
	\begin{minipage}{0.43\textwidth}
		\centering
		\includegraphics[height=2.7cm,width=4.3cm]{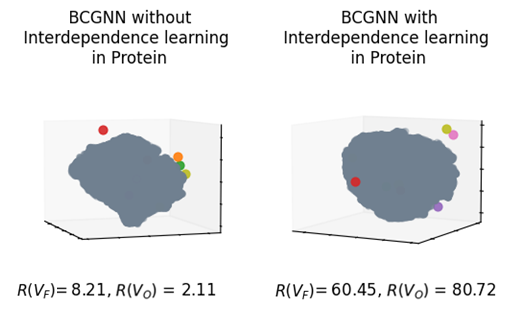}
	\end{minipage}
	\vspace{-3pt}
	\caption{\textbf{Left}: Average test MAE of feature imputation under MCAR, MAR and MNAR with different missing rates in Cancer, Wine, Power, Energy, Protein over 5 random trials. \textbf{Right}: The embedding spaces $V_F$ and $V_O$ for feature and observation nodes, respectively. They are obtained from the trained BCGNN with/without learning interdependence structure in the above datasets under MAR with a missing rate of 0.3. The \textbf{colored} dots represent feature node embeddings and the \textbf{grey} dots represent observation node embeddings.
	} 
	\label{featureimputationgraph}
	\vspace{-4mm}
\end{figure*}

\section{Additional Experimental Results}
\subsection{Additional Experimental Results for Feature Imputation}
\label{comprehensive}
Figure \ref{featureimputationMAR} reports the average test MAE of BCGNN and baseline models in UCI datasets under MAR with a missing rate of 0.3 over 5 random trials, and Figure \ref{featureimputationgraph} respectively reports the average test MAE of BCGNN and baseline models on the Wine, Cancer, Power, Energy and Protein datasets under different missing mechanisms  and different missing rate over 5 random trials.

\subsection{Experimental Results for  Generalization on Unseen Data and Label Prediction}
\label{labelpredictionandgeneralization}
\begin{figure}[h]
	\centering
	\begin{minipage}{0.50\textwidth}
		\centering
		\includegraphics[height=4cm,width=6.9cm]{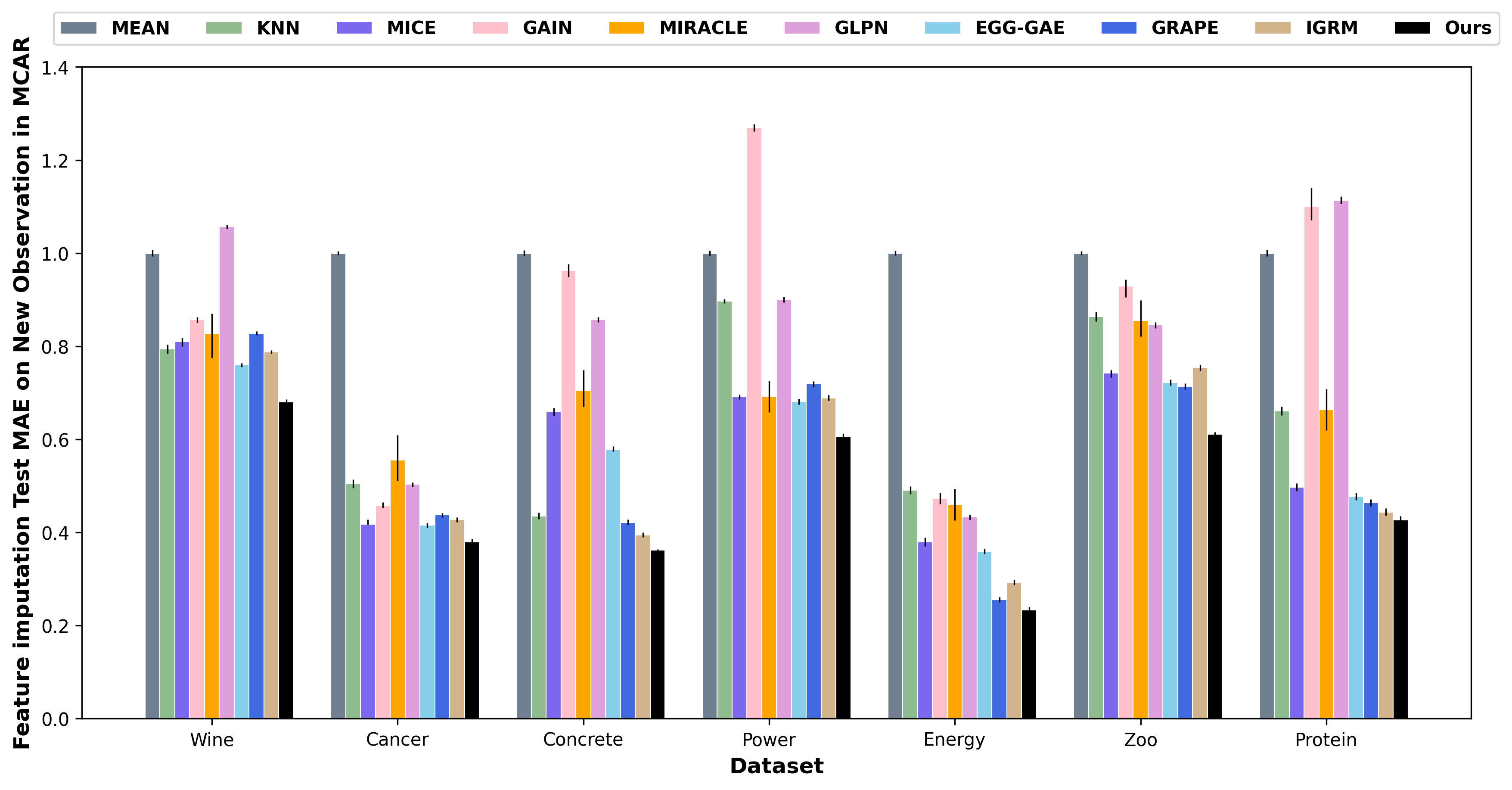}
	\end{minipage}%
	\hfill
	\begin{minipage}{0.48\textwidth}
		\centering
		\includegraphics[height=4cm,width=6.9cm]{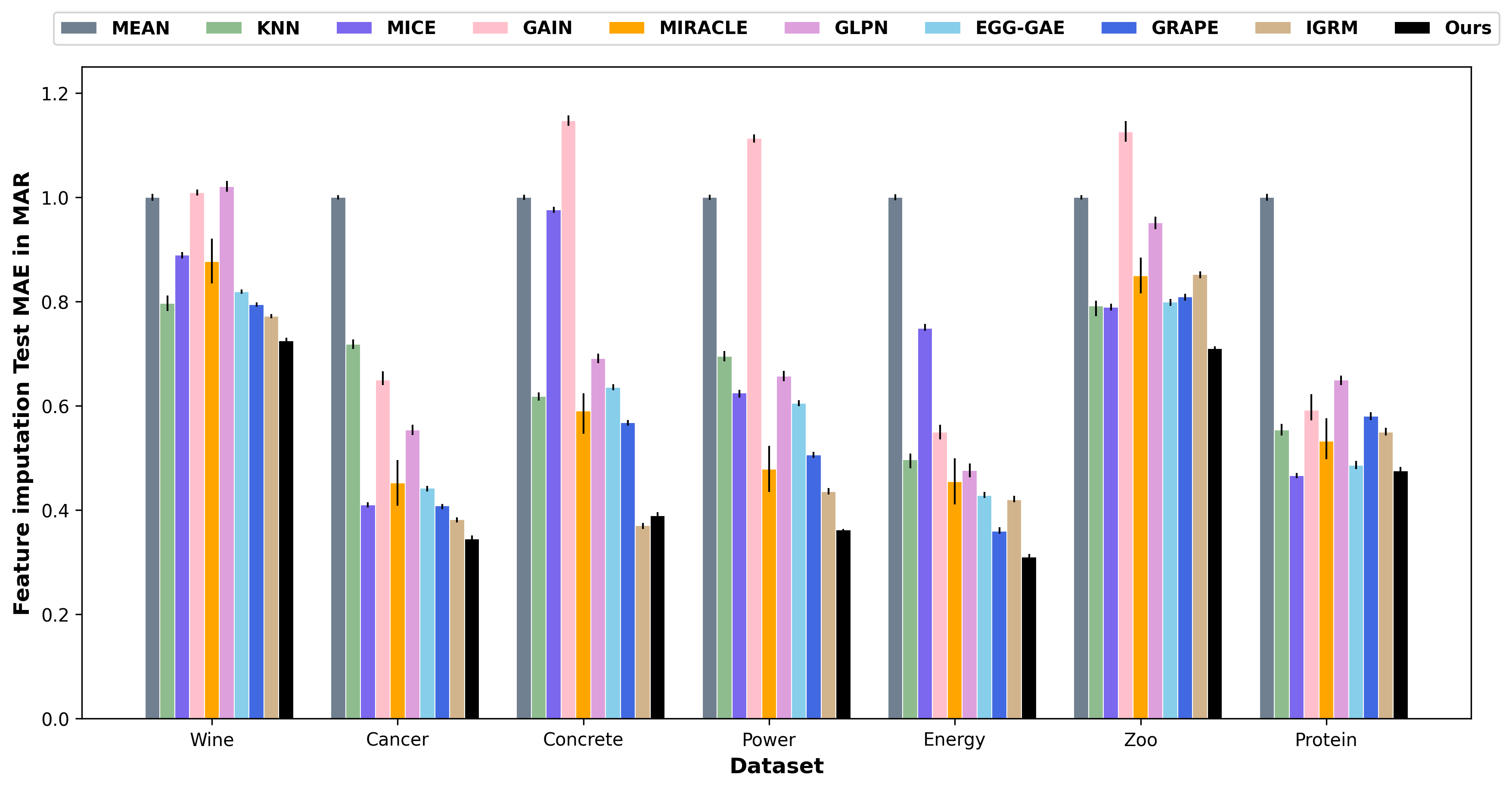}
	\end{minipage}
	\vspace{3pt} 
	
	\begin{minipage}{1\textwidth}
		\centering
		\includegraphics[height=4cm,width=7cm]{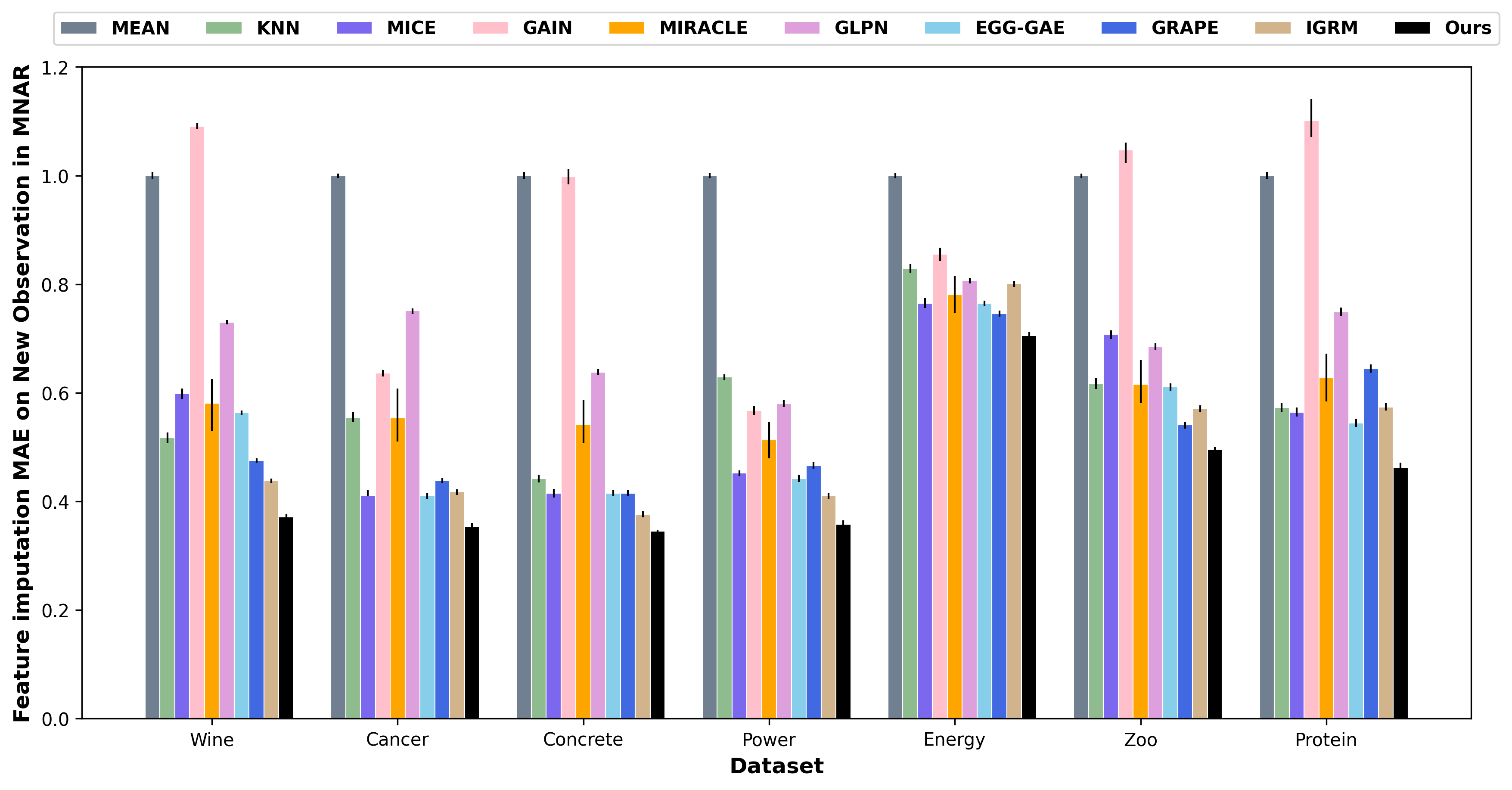}
	\end{minipage}%
	
	\caption{Average MAE of feature imputation on new observations at a missing rate of 0.3  under MCAR, MAR and MNAR in UCI datasets over 5 random trials. The result is normalized by the average performance of the Mean imputation. Compared with the best baseline (GRAPE), BCGNN demonstrates reductions in the average MAE of 12.3\%, 14.8\%, and 18.2\% under MCAR, MAR and MNAR, respectively.} 
	\label{generalization}
	\vspace{-3mm}
\end{figure}

At a missing rate of 0.3, we test the performance of BCGNN and baseline models in feature imputation tasks when handling new observations with missing data under different missing mechanisms. We report the average MAE in Figure \ref{generalization}. 

\begin{figure}[h]
	\centering
	\begin{minipage}{0.50\textwidth}
		\centering
		\includegraphics[height=4cm,width=6.8cm]{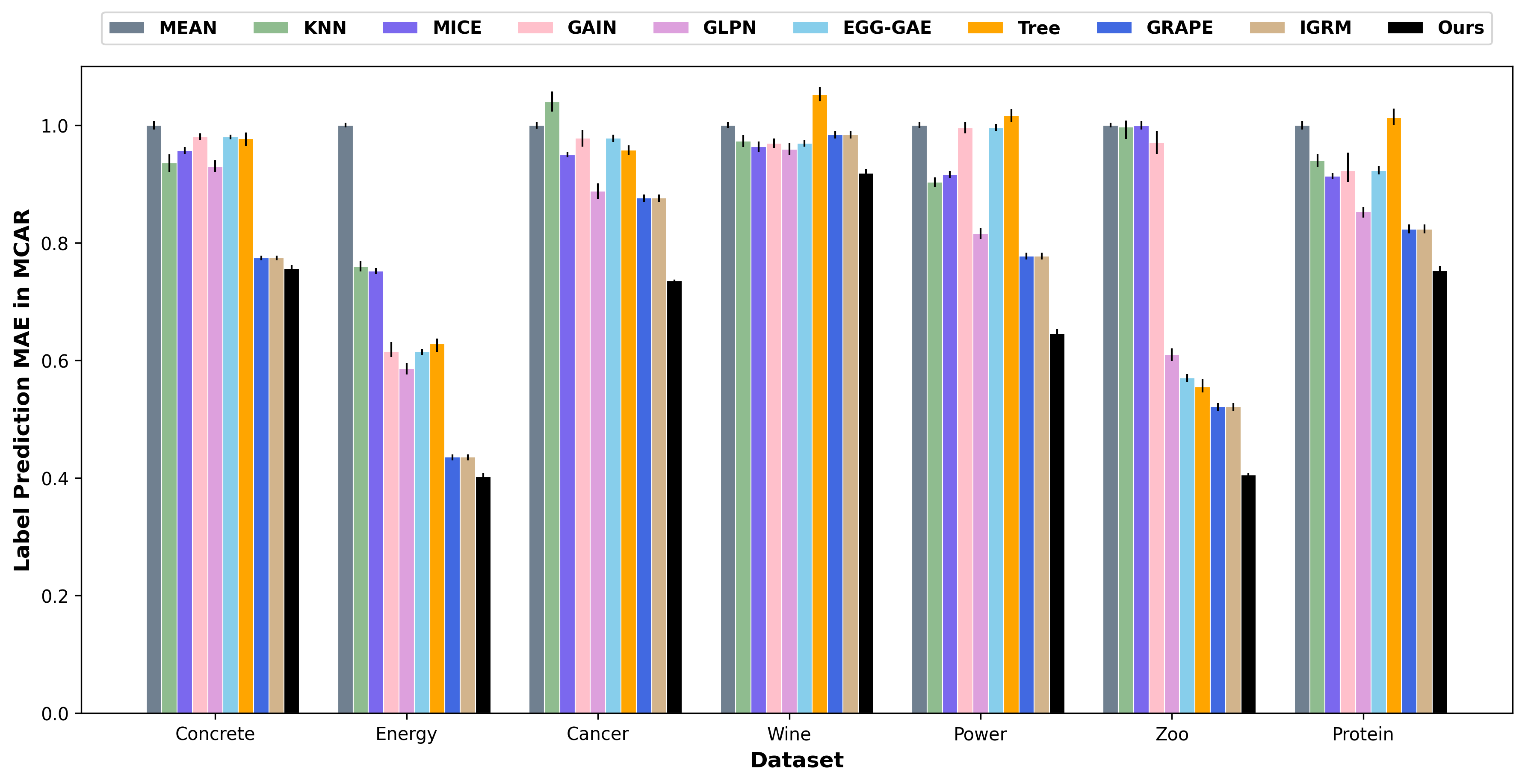}
	\end{minipage}%
	\hfill
	\begin{minipage}{0.48\textwidth}
		\centering
		\includegraphics[height=4cm,width=6.8cm]{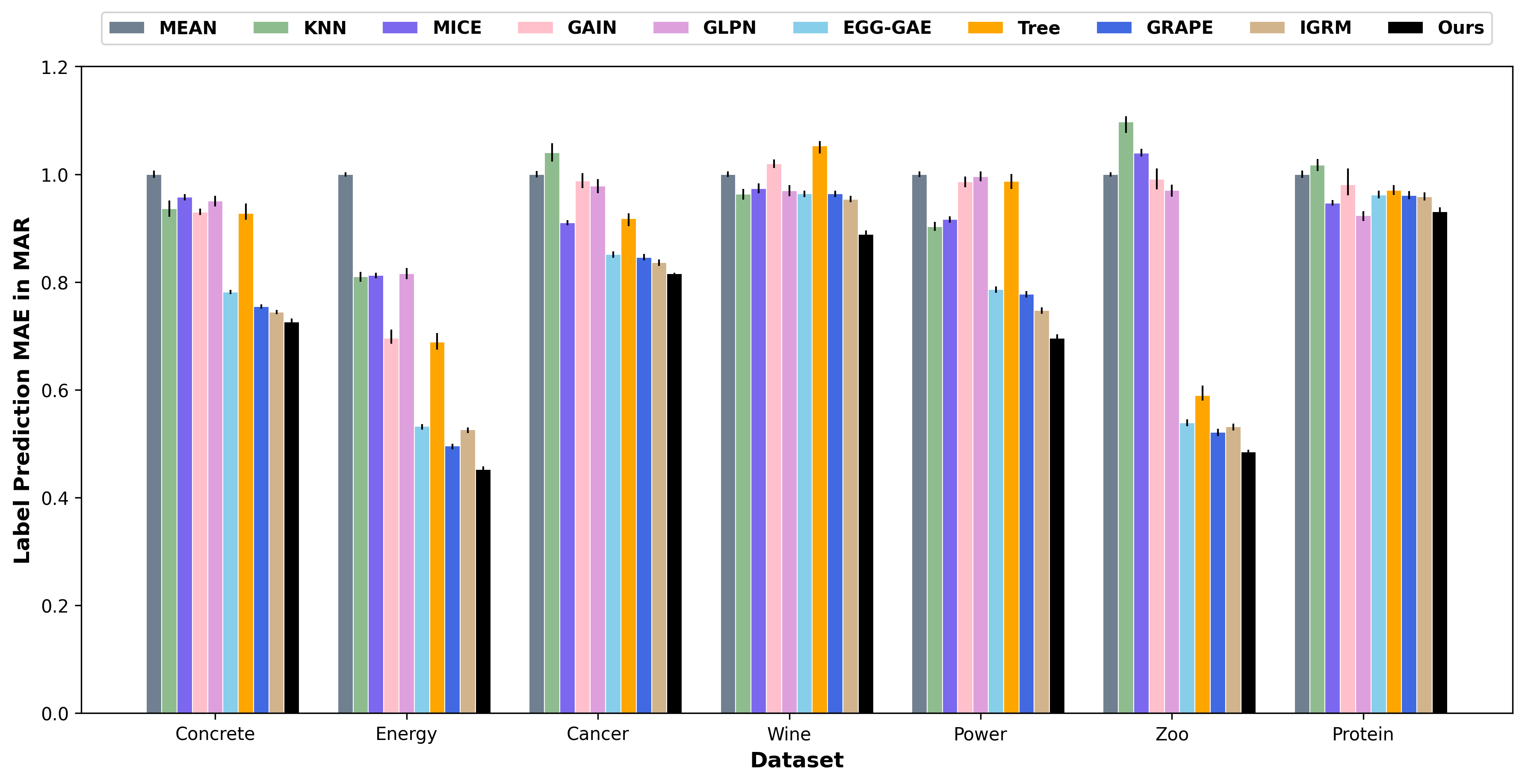}
	\end{minipage}
	\vspace{3pt} 
	
	\begin{minipage}{1\textwidth}
		\centering
		\includegraphics[height=4cm,width=7cm]{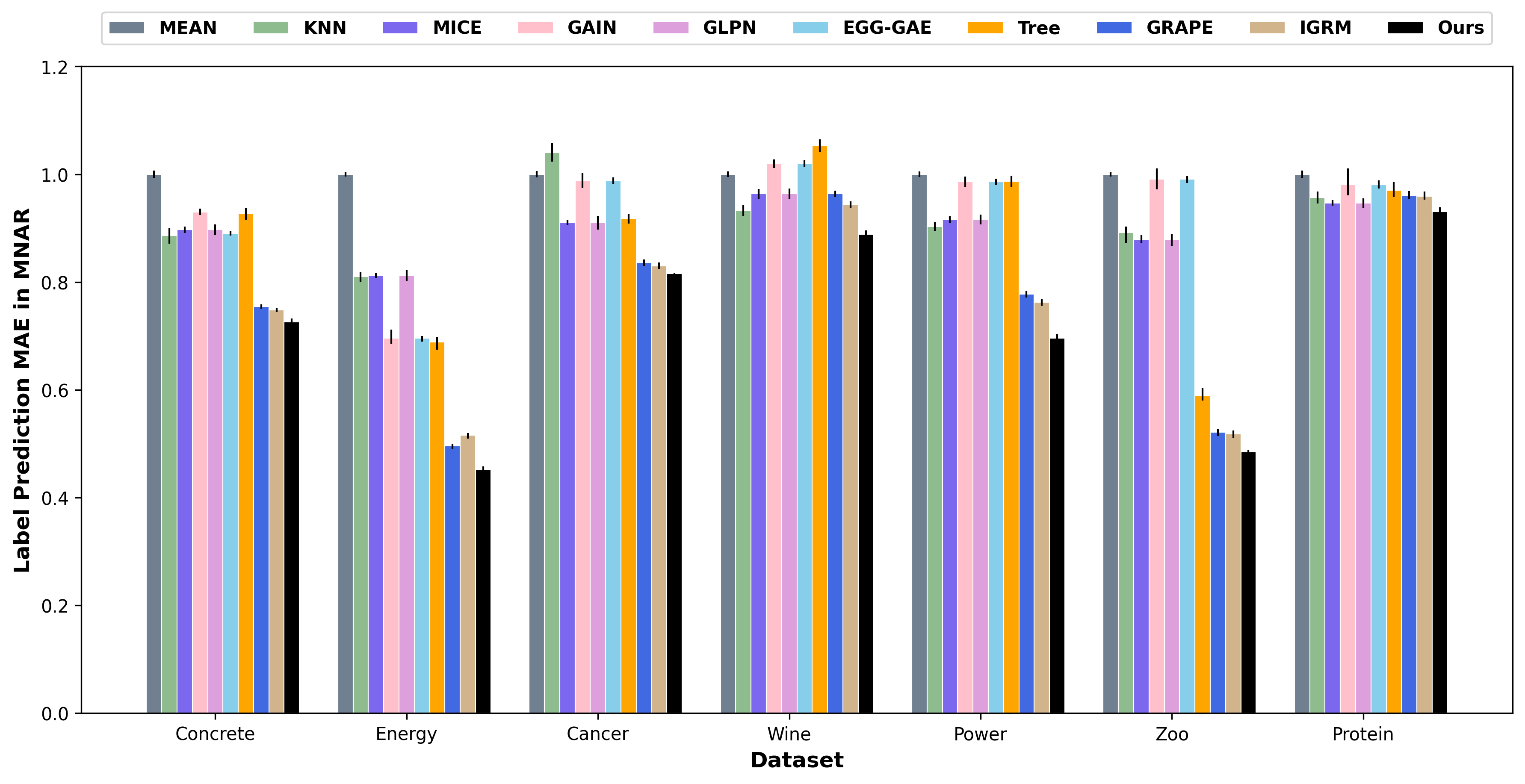}
	\end{minipage}%
	
	\caption{Average MAE of label prediction at a  missing rate of 0.3 under MCAR,MAR and MNAR in UCI datasets over 5 random trials. The results are normalized by the average performance of the Mean imputation.}
	\label{Labelprediction}
\end{figure}

At a  missing rate of 0.3, we evaluate  the performance of BCGNN and baseline models in the label prediction task with missing data under  various missing mechanisms. We present the average MAE for label prediction in Figure \ref{Labelprediction}.

\begin{figure}[t]
	\centering
	\begin{minipage}{0.45\textwidth}
		\centering
		\includegraphics[height=4cm,width=7cm]{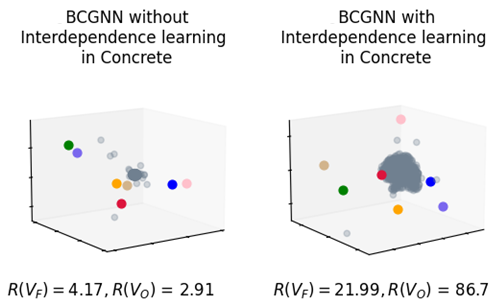}
	\end{minipage}%
	\hfill
	\begin{minipage}{0.45\textwidth}
		\centering
		\includegraphics[height=4cm,width=7cm]{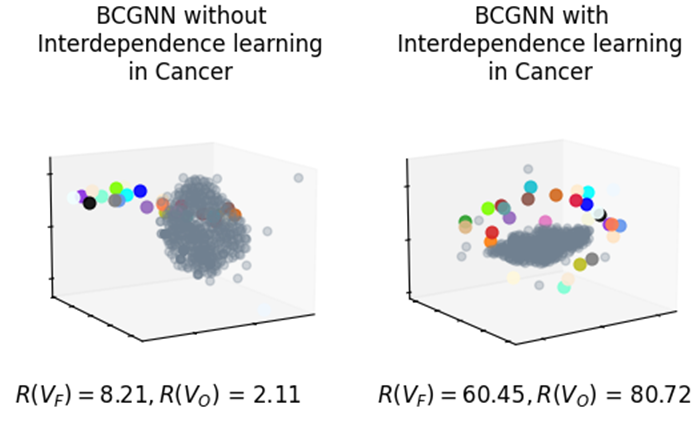}
	\end{minipage}
	\vspace{5pt} 
	\begin{minipage}{0.55\textwidth}
		\centering
		\includegraphics[height=4cm,width=7cm]{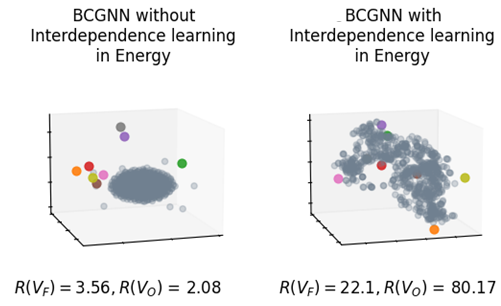}
	\end{minipage}%
	\begin{minipage}{0.55\textwidth}
		\centering
		\includegraphics[height=4cm,width=7cm]{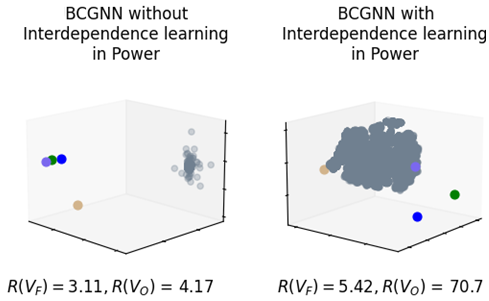}
	\end{minipage}%
	\vspace{5pt} 
	\begin{minipage}{0.55\textwidth}
		\centering
		\includegraphics[height=4cm,width=7cm]{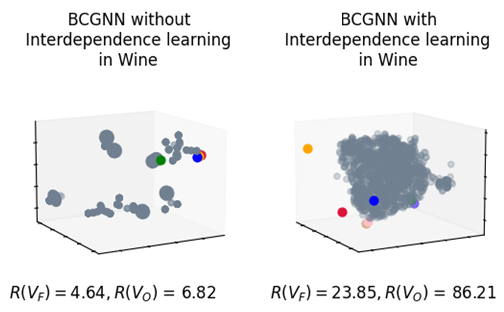}
	\end{minipage}%
	\begin{minipage}{0.55\textwidth}
		\centering
		\includegraphics[height=3.9cm,width=6.8cm]{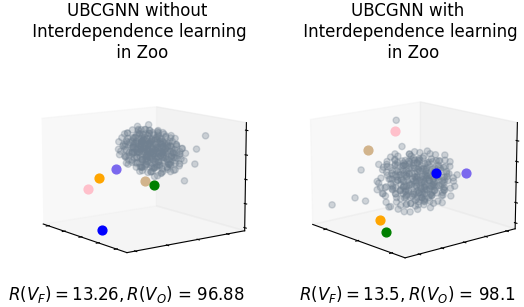}
	\end{minipage}%
	\vspace{5pt}
	\begin{minipage}{1\textwidth}
		\centering
		\includegraphics[height=4cm,width=7cm]{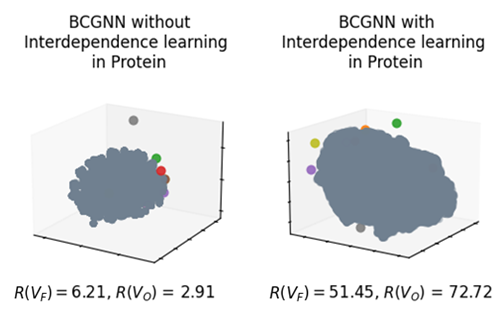}
	\end{minipage}%
	\caption{The sizes of feature embedding space $V_F$ and observation embedding space $V_O$, which are obtained from the trained BCGNN with/without interdependence structure learning under MCAR by the dimension reduction visualization method t-SNE for all datasets.}
	\label{TSNEMCAR}
\end{figure}

\begin{figure}[h]
	\centering
	\begin{minipage}{0.45\textwidth}
		\centering
		\includegraphics[height=4cm,width=7cm]{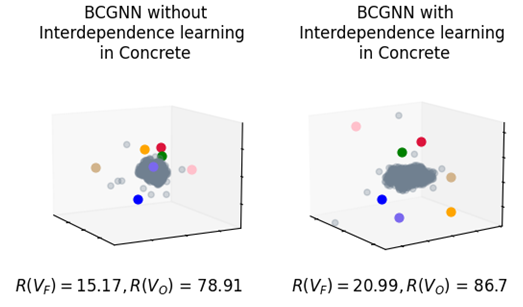}
	\end{minipage}%
	\hfill
	\begin{minipage}{0.45\textwidth}
		\centering
		\includegraphics[height=4cm,width=7cm]{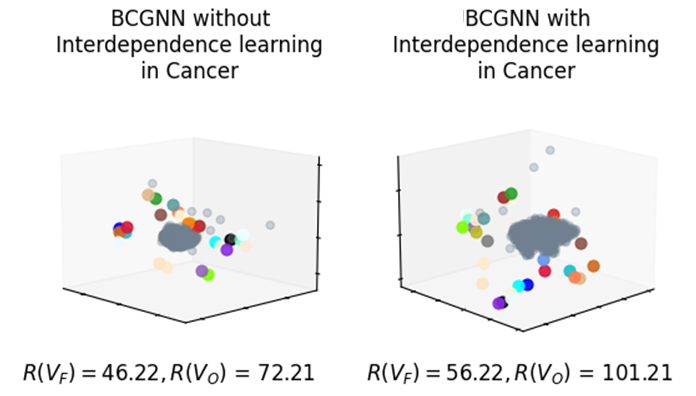}
	\end{minipage}
	\vspace{5pt} 
	\begin{minipage}{0.55\textwidth}
		\centering
		\includegraphics[height=4cm,width=7cm]{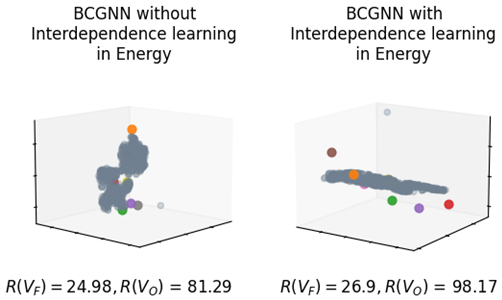}
	\end{minipage}%
	\begin{minipage}{0.55\textwidth}
		\centering
		\includegraphics[height=4cm,width=7cm]{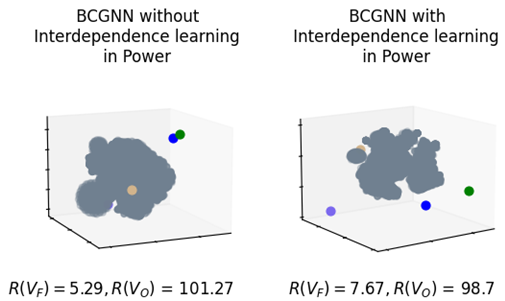}
	\end{minipage}%
	\vspace{5pt} 
	\begin{minipage}{0.55\textwidth}
		\centering
		\includegraphics[height=4cm,width=7cm]{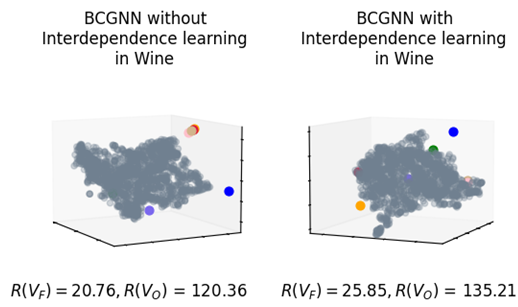}
	\end{minipage}%
	\begin{minipage}{0.55\textwidth}
		\centering
		\includegraphics[height=3.9cm,width=6.8cm]{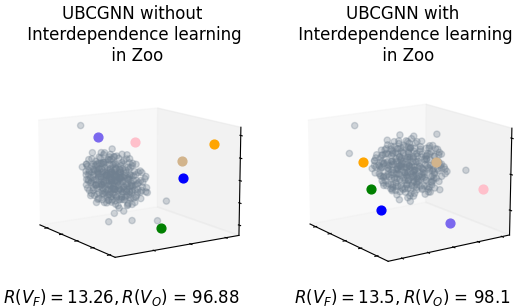}
	\end{minipage}%
	\vspace{5pt}
	\begin{minipage}{1\textwidth}
		\centering
		\includegraphics[height=4cm,width=7cm]{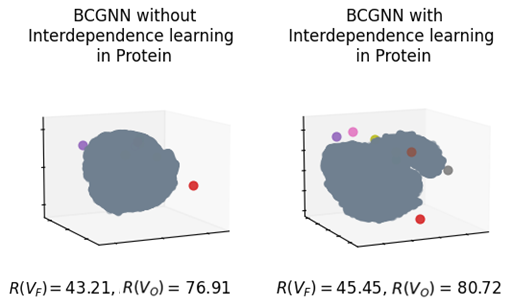}
	\end{minipage}%
	\caption{The sizes of feature embedding space $V_F$ and observation embedding space $V_O$, which are obtained from the trained BCGNN with/without interdependence structure learning under MNAR by the dimension reduction visualization method t-SNE for all datasets.}
	\label{TSNEMNAR}
\end{figure}

\vspace{-5pt}

\subsection{Ablation Study on Aggregation Function and Correlation Estimates}
\label{ablation}
\textbf{Aggregation Function.} In BCGNN, to perform message aggregation on both feature nodes and observation nodes, different aggregation functions such as MEAN($\cdot$), SUM($\cdot$), and MAX($\cdot$) can be used. We evaluate the impact of different aggregation functions on the performance of BCGNN by replicating the experiments in Section \ref{featureimputation} under MCAR with a missing rate of~0.3.

\setlength{\tabcolsep}{4pt} 
\begin{table}[h]
	\centering
	\caption{The average test MAE for BCGNN with different aggregation functions and correlation coefficients}
		\begin{tabular}{cccccccc}
			\toprule
			\textbf{} & \textbf{Concrete} & \textbf{Cancer} & \textbf{Energy} & \textbf{Protein} & \textbf{Wine}  & \textbf{Zoo}  & \textbf{Power}\\ 
			\midrule
			BCGNN with \textbf{SUM($\cdot$)} & 0.0732 & \textbf{0.0422} & \textbf{0.0701} &0.0318 & 0.0683 & 0.1483 & 0.1023 \\
			BCGNN with \textbf{MEAN($\cdot$)} & \textbf{0.0731} & 0.0423 & 0.0706 &\textbf{0.0310} & \textbf{0.0678} & \textbf{0.1452} & \textbf{0.1012} \\
			BCGNN with \textbf{MAX($\cdot$)} & 0.0753 & 0.0451 & 0.0752 & 0.0351 & 0.0701 & 0.1501 & 0.1085  \\
			\midrule
			BCGNN with \textbf{Spearman} & \textbf{0.0731} & \textbf{0.0423} & \textbf{0.0706} &\textbf{0.0310} & \textbf{0.0678}  & \textbf{0.1452} & \textbf{0.1012} \\
			BCGNN with \textbf{Pearson} & 0.0770 & 0.0454 & 0.0724 & 0.0325 & 0.0688 & 0.1480  & 0.1072    \\
			BCGNN with \textbf{Kendall} & 0.0755 & 0.0441 & 0.0715 & 0.0321 & 0.0681 & 0.1511  & 0.1041    \\
			\bottomrule
		\end{tabular}
	\label{Aggregate}
\end{table}

From Table \ref{Aggregate}, it is evident that BCGNN performs best with the SUM($\cdot$) function, while it particularly underperforms with MAX($\cdot$). The reason for this is that employing MAX($\cdot$) as the aggregation function in the message aggregation process for feature nodes may lead to an overemphasis on messages from neighboring feature nodes (due to their tendency to be stronger than those from observation nodes), thereby overshadowing the messages received from observation nodes. This unbalanced choice reduces the expressive capacity of feature nodes.

\textbf{Correlation Estimators.} 
We investigate the influence of various correlation estimators on the performance of BCGNN. Specifically, we utilize Spearman, Pearson, and Kendall correlation coefficients in the model, respectively. Table \ref{Aggregate} indicates that the choice of correlation estimator truly impacts BCGNN's performance. Spearman and Kendall correlation coefficients generally demonstrate superior performance,  whereas the Pearson correlation coefficient shows relatively weaker results. This discrepancy arises from the fact that the Pearson correlation coefficient is primarily suitable for continuous data and is constrained to capturing linear relationships exclusively. Conversely, both Kendall and Spearman correlation coefficients are computed based on ranks, rendering them nonparametric methods that do not require distributional assumptions and are adept at capturing nonlinear relationships.

\subsection{Ablation Study on DropEdge and AttentionDrop Scheme}
\label{Dropedge}
We evaluate the performance of BCGNN under three missing mechanisms with a missing rate of 0.3, which utilizes DropEdge on both bipartite graph and complete directed graph, as well as AttentionDrop on the complete directed graph. We evaluate the BCGNN variants without one, two, or three of these schemes. We report the averaged MAE for feature imputation under different scenarios in Table \ref{B}.

\setlength{\tabcolsep}{2pt} 
\begin{table}[h]
	\small
	\centering
	\caption{Average test MAE of feature imputation tasks for BCGNN with or without DropEdge and AttentionDrop schemes}
	
		\begin{tabular}{cccccccc}
			\toprule
			\textbf{MCAR} & \textbf{Concrete} & \textbf{Cancer} & \textbf{Energy} & \textbf{Zoo} & \textbf{Wine} & \textbf{Power} &\textbf{Protein} \\ 
			\midrule
			\textbf{Drop in $\mathcal{G}_B$}$\times$ \ \ \textbf{Drop in $\mathcal{G}_C$}$\times$ \ \ \textbf{AttDrop}$\times$ & 0.142 & 0.072 & 0.120 & 0.171 & 0.109 & 0.152 & 0.061 \\ 
			\textbf{Drop in $\mathcal{G}_B$}\checkmark \ \ \textbf{Drop in $\mathcal{G}_C$}$\times$ \ \ \textbf{AttDrop}$\times$ & 0.078 & 0.046 & 0.075 & 0.149 & 0.070 & 0.105 & 0.032 \\ 
			\textbf{Drop in $\mathcal{G}_B$}\checkmark \ \ \textbf{Drop in $\mathcal{G}_C$}\checkmark \ \ \textbf{AttDrop}$\times$ & 0.075 & 0.044 & 0.072 & 0.146 & 0.066 & 0.103 & 0.030 \\
			\textbf{Drop in $\mathcal{G}_B$}\checkmark \ \ \textbf{Drop in $\mathcal{G}_C$}\checkmark \ \ \textbf{AttDrop}\checkmark & \textbf{0.073} & \textbf{0.042} & \textbf{0.070} & \textbf{0.145} & \textbf{0.066} & \textbf{0.101} & \textbf{0.030} \\
			\bottomrule
		\end{tabular}%
		\begin{tabular}{cccccccc}
			\toprule
			\textbf{MAR} & \textbf{Concrete} & \textbf{Cancer} & \textbf{Energy} & \textbf{Zoo} & \textbf{Wine} & \textbf{Power} &\textbf{Protein} \\
			\midrule
			\textbf{Drop in $\mathcal{G}_B$}$\times$ \ \ \textbf{Drop in $\mathcal{G}_C$}$\times$ \ \ \textbf{AttDrop}$\times$ & 0.073 & 0.072 & 0.143 & 0.169 & 0.073 & 0.132 & 0.071\\
			\textbf{Drop in $\mathcal{G}_B$}\checkmark \ \ \textbf{Drop in $\mathcal{G}_C$}$\times$ \ \ \textbf{AttDrop}$\times$ & 0.053 & 0.054 & 0.119 & 0.149 & 0.061 & 0.102 & 0.052 \\
			\textbf{Drop in $\mathcal{G}_B$}\checkmark \ \ \textbf{Drop in $\mathcal{G}_C$}\checkmark \ \ \textbf{AttDrop}$\times$ & 0.051 & 0.051 & 0.115 & 0.147 & 0.059 & 0.098 & 0.049 \\
			\textbf{Drop in $\mathcal{G}_B$}\checkmark \ \ \textbf{Drop in $\mathcal{G}_C$}\checkmark \ \ \textbf{AttDrop}\checkmark & \textbf{0.049} & \textbf{0.050} & \textbf{0.115} & \textbf{0.145} & \textbf{0.057} & \textbf{0.096} & \textbf{0.048}\\
			\bottomrule
		\end{tabular}%
		\begin{tabular}{cccccccc}
			\toprule
			\textbf{MNAR} & \textbf{Concrete} & \textbf{Cancer} & \textbf{Energy} & \textbf{Zoo} & \textbf{Wine} & \textbf{Power} &\textbf{Protein}\\
			\midrule
			\textbf{Drop in $\mathcal{G}_B$}$\times$ \ \ \textbf{Drop in $\mathcal{G}_C$}$\times$ \ \ \textbf{AttDrop}$\times$ & 0.072 & 0.073 & 0.123 & 0.161 & 0.081 & 0.122 & 0.072 \\
			\textbf{Drop in $\mathcal{G}_B$}\checkmark \ \ \textbf{Drop in $\mathcal{G}_C$}$\times$ \ \ \textbf{AttDrop}$\times$ & 0.059 & 0.055 & 0.100 & 0.145 & 0.066 & 0.102 & 0.052 \\
			\textbf{Drop in $\mathcal{G}_B$}\checkmark \ \ \textbf{Drop in $\mathcal{G}_C$}\checkmark \ \ \textbf{AttDrop}$\times$ & 0.056 & 0.053 & 0.096 & 0.142 & 0.063 & 0.099  & 0.048\\
			\textbf{Drop in $\mathcal{G}_B$}\checkmark \ \ \textbf{Drop in $\mathcal{G}_C$}\checkmark \ \ \textbf{AttDrop}\checkmark & \textbf{0.054} & \textbf{0.052} & \textbf{0.094} & \textbf{0.141} & \textbf{0.061} & \textbf{0.097} &\textbf{0.046} \\
			\bottomrule
		\end{tabular}%
	\vspace{1mm}
	\label{B}
\end{table}

\setlength{\tabcolsep}{6pt} 
\begin{table}[h]
	\centering
	\caption{Average running  time (in seconds) for feature imputation by different methods}
	\begin{tabular}{cccccccc}
		\hline
		Model      & Concrete  & Energy   & Cancer & Zoo  & Protein & Wine   & Power  \\ \hline
		Mean       & 0.0008  & 0.0010 & 0.0010 & 0.0008 & 0.0135 & 0.0011 & 0.0015 \\ 
		KNN        & 0.2251    & 0.1343   & 0.1720  & 0.0278  & 636.28    & 0.5041   &11.459    \\ 
		MICE       & 0.0310    & 0.0321   & 0.3720  & 0.0281  & 0.2716  & 0.0531  & 0.0291  \\ 
		SVD        & 0.0691    & 0.0202   & 0.2710  & 0.0432  & 0.5932  & 0.0564  &0.1427   \\ 
		Spectral   & 0.0710    & 0.0583   & 0.0830  & 0.0341  & 1.4333 & 0.0978  & 0.1814   \\ 
		GAIN       & 0.1202    & 0.0132   & 0.0172  & 0.0121  & 0.0501 & 0.0131  & 0.0146  \\ 
		MIRACLE    & 0.0912    & 0.0928   & 0.1210  & 0.8711  & 2.4213  & 0.0872  & 0.0871  \\ 
		GLPN       & 0.1821    & 0.1921   & 0.2121  & 0.1621  & 0.5213  & 0.1732  & 0.1853  \\ 
		GRAPE      & 0.0273    & 0.0151   & 0.0331  & 0.0133  & 0.5683  & 0.0199  & 0.0488  \\ 
		BCGNN      & 0.0403    & 0.0232   & 0.1052  & 0.0231  & 0.8323  & 0.0253  & 0.0572  \\
		\hline
	\end{tabular}
	\label{runningtime0}
\end{table}

\subsection{Running Time Comparison}
We present the running times associated with feature imputation using various methods in our analysis. Specifically, for methods such as Mean, KNN, MICE, SVD, and Spectral, the reported running time corresponds to the duration of a single function call dedicated to imputing missing data. Conversely, for models like GAIN, MIRACLE, GLPN, GRAPE, and our BCGNN, the running time is defined as the duration of one forward pass through the model. Table \ref{runningtime0} shows the average running time over 5 random trials for the experiments in Section \ref{featureimputation}.

\section{Computational Hardware}
\label{hardware}
All models are trained on a Windows 10 64-bit OS (version 19045) with 16GB of RAM (AMD Ryzen 7 4800H CPU @ 2.9GHz) and 2 NVIDIA GeForce RTX 2060 with Max-Q Design GPUs.

\section{Broader Impacts and Limitations}
Our method demonstrates superior performance in both feature imputation and label prediction tasks by leveraging
the complex interdependence among features for tabular data. It enhances
the applicability of missing data imputation tasks to real-world problems, such as clinical research, finance, economics, and microarray analysis. Ethically, we believe that our rather
fundamental work has minimal potential for misuse. One limitation is our inability to effectively integrate the causal structure among feature nodes to improve feature imputation accuracy. This remains a significant area of research interest for us in the future.

\end{appendices}

\clearpage
\bibliography{sn-bibliography}

\end{document}